\documentclass{article}
\newcount\Comments  % 1 suppresses notes to selves in text
\usepackage{color}
\definecolor{red}{rgb}{1,0,0}
\definecolor{darkgreen}{rgb}{0,0.5,0}
\definecolor{darkblue}{rgb}{0,0,0.5}
\definecolor{purple}{rgb}{1,0,1}
\newcommand{\kibitz}[2]{\ifnum\Comments=0\textcolor{#1}{#2}\fi}

\newcommand{\first}{\includegraphics[height=0.9em]{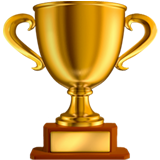}\ }
\newcommand{\second}{\includegraphics[height=0.9em]{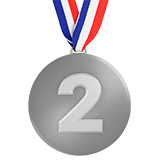}}
\newcommand{\third}{\includegraphics[height=0.9em]{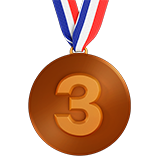}}
\usepackage{threeparttable}
\usepackage{microtype}
\usepackage{graphicx}
\usepackage{tablefootnote}
\usepackage{subfigure}
\usepackage{booktabs}
\usepackage{multirow}
\usepackage{hyperref}
\usepackage[table]{xcolor}
\usepackage{multirow}
\usepackage{hhline}
\usepackage{colortbl}

\definecolor{lightgray}{gray}{0.9}

\newcommand{\diff}[1]{\textcolor{red}{#1}} % Command to highlight differences

\usepackage[accepted]{icml2023}

\usepackage{amsmath}
\usepackage{amssymb}
\usepackage{mathtools}
\usepackage{amsthm}

\usepackage[capitalize,noabbrev]{cleveref}

\theoremstyle{plain}

\theoremstyle{definition}

\theoremstyle{remark}

\hypersetup{
	colorlinks=true,
	urlcolor=magenta,
}
\usepackage[textsize=tiny]{todonotes}

\icmltitlerunning{Benchmarking Large Multimodal Models against Common Corruptions}

\begin{document}

\twocolumn[
\icmltitle{Benchmarking Large Multimodal Models against Common Corruptions}



\icmlsetsymbol{claim}{*}
\icmlsetsymbol{claim2}{$\dagger$}

\begin{icmlauthorlist}
\icmlauthor{Jiawei Zhang}{claim,to1}
\icmlauthor{Tianyu Pang}{to2}
\icmlauthor{Chao Du}{to2}
\icmlauthor{Yi Ren}{claim2,to3}
\icmlauthor{Bo Li}{to1,to4}
\icmlauthor{Min Lin}{to2}
\end{icmlauthorlist}

\icmlaffiliation{to1}{UIUC}
\icmlaffiliation{to2}{Sea AI Lab}
\icmlaffiliation{to3}{ByteDance}
\icmlaffiliation{to4}{University of Chicago}
% \icmlaffiliation{to3}{School of ZZZ, Institute of WWW, Location, Country}
\icmlcorrespondingauthor{Jiawei Zhang}{jiaweiz7@illinois.edu}
\icmlcorrespondingauthor{Tianyu Pang}{tianyupang@sea.com}
\icmlcorrespondingauthor{Chao Du}{duchao@sea.com}
% \icmlcorrespondingauthor{Bo Li}{lbo@illinois.edu}

\icmlkeywords{Machine Learning, ICML}

\vskip 0.3in
]



\printAffiliationsAndNotice{$^{*}$Work done during an associate membership at Sea AI Lab. $^\dagger$Work done while at Sea AI Lab.} % otherwise use the standard text.

\begin{abstract}
% This paper addresses a critical gap in the evaluation of multimodal models, focusing on the self-consistency of outputs under common corruption scenarios. We explore the cross-modality interactions involving text, image, and speech, covering four fundamental tasks: text to image, image to text, speech to text, and text to speech. Our contributions include the development of a comprehensive benchmark for assessing self-consistency in multimodal models, a novel methodology for selecting representative examples from large datasets, and the introduction of a consistent metric system for performance measurement across various cross-modalities. We employ this framework to conduct extensive analyses of state-of-the-art multimodal models, revealing xxx. The project is available at~\url{https://MMCBench.github.io}\jiawei{add placeholder for repo here}.
This technical report aims to fill a deficiency in the assessment of large multimodal models (LMMs) by specifically examining the self-consistency of their outputs when subjected to common corruptions. We investigate the cross-modal interactions between text, image, and speech, encompassing four essential generation tasks: \emph{text-to-image}, \emph{image-to-text}, \emph{text-to-speech}, and \emph{speech-to-text}. We create a comprehensive benchmark, named \textbf{MMCBench}, that covers 
% numerous \chao{use detail numbers}\jiawei{one question is that we mainly use LMM for image to text tasks, while for other tasks, the models are not LMM} 
\textbf{more than 100}
popular LMMs (totally \textbf{over 150} model checkpoints).
% Our benchmarking pipeline involves selecting representative examples from large datasets and implementing a unified metric system to measure performance across different cross-modalities.
A thorough evaluation under common corruptions is critical for practical deployment and facilitates a better understanding of the reliability of cutting-edge LMMs. The benchmarking code is available at \href{https://github.com/sail-sg/MMCBench}{https://github.com/sail-sg/MMCBench}.\looseness=-1 
%\jiawei{add how we get the abbreviation for MMCBench later}
\end{abstract}

\begin{figure*}[th] 
  \centering
  \includegraphics[width=\textwidth]{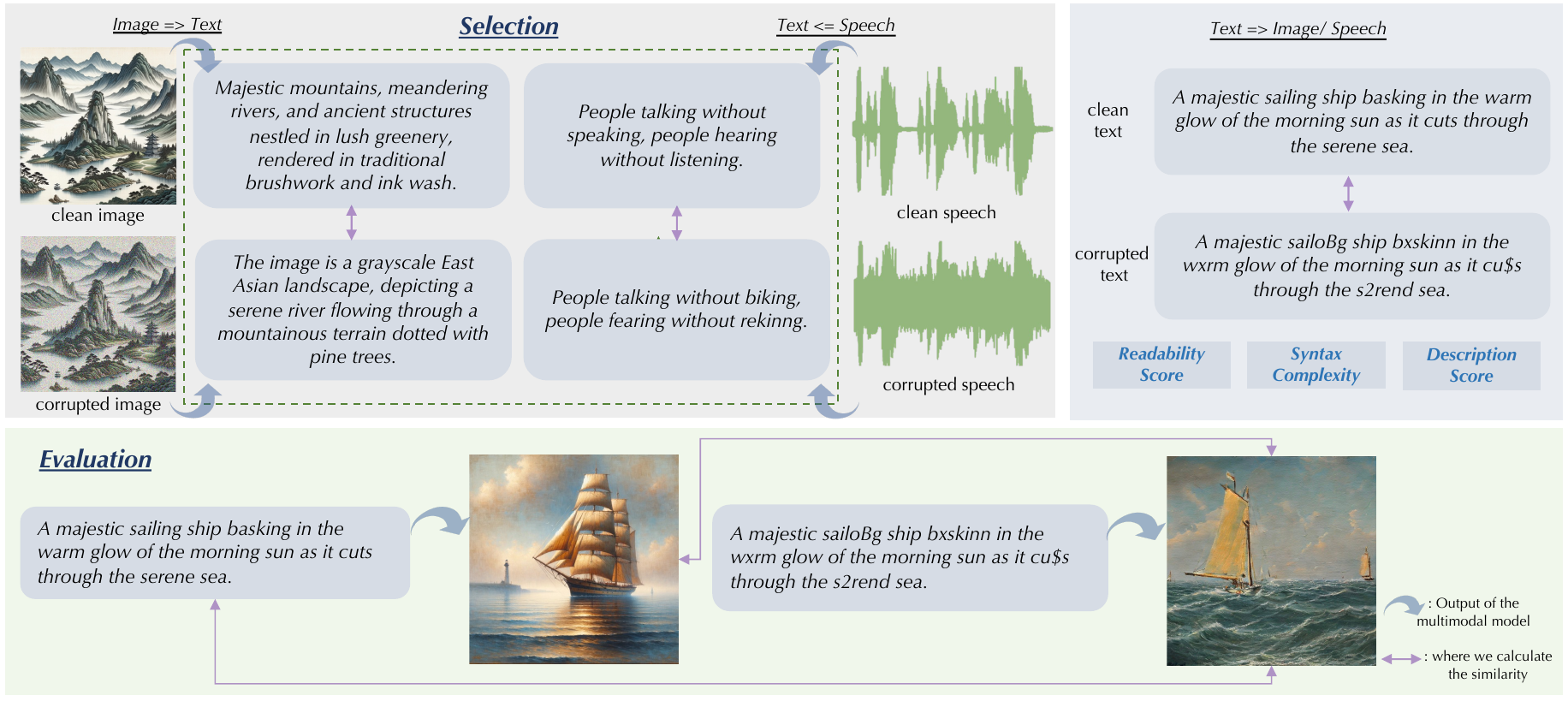}
  \vspace{-6mm}
  \caption{Overview of the selection and evaluation process for cross-modality consistency. The selection process (\emph{top row}) involves determining similarity based on text modality, using model-generated captions or transcriptions for non-text inputs, while directly comparing text inputs before and after corruption. The evaluation process (\emph{bottom row}) measures self-consistency by comparing clean inputs with outputs from corrupted inputs and by comparing outputs from clean and corrupted inputs against each other.}
  \label{fig:pipeline}
  \vspace{3mm}
\end{figure*}

\begin{figure*}[th] 
  \centering
  \includegraphics[width=0.9\textwidth]{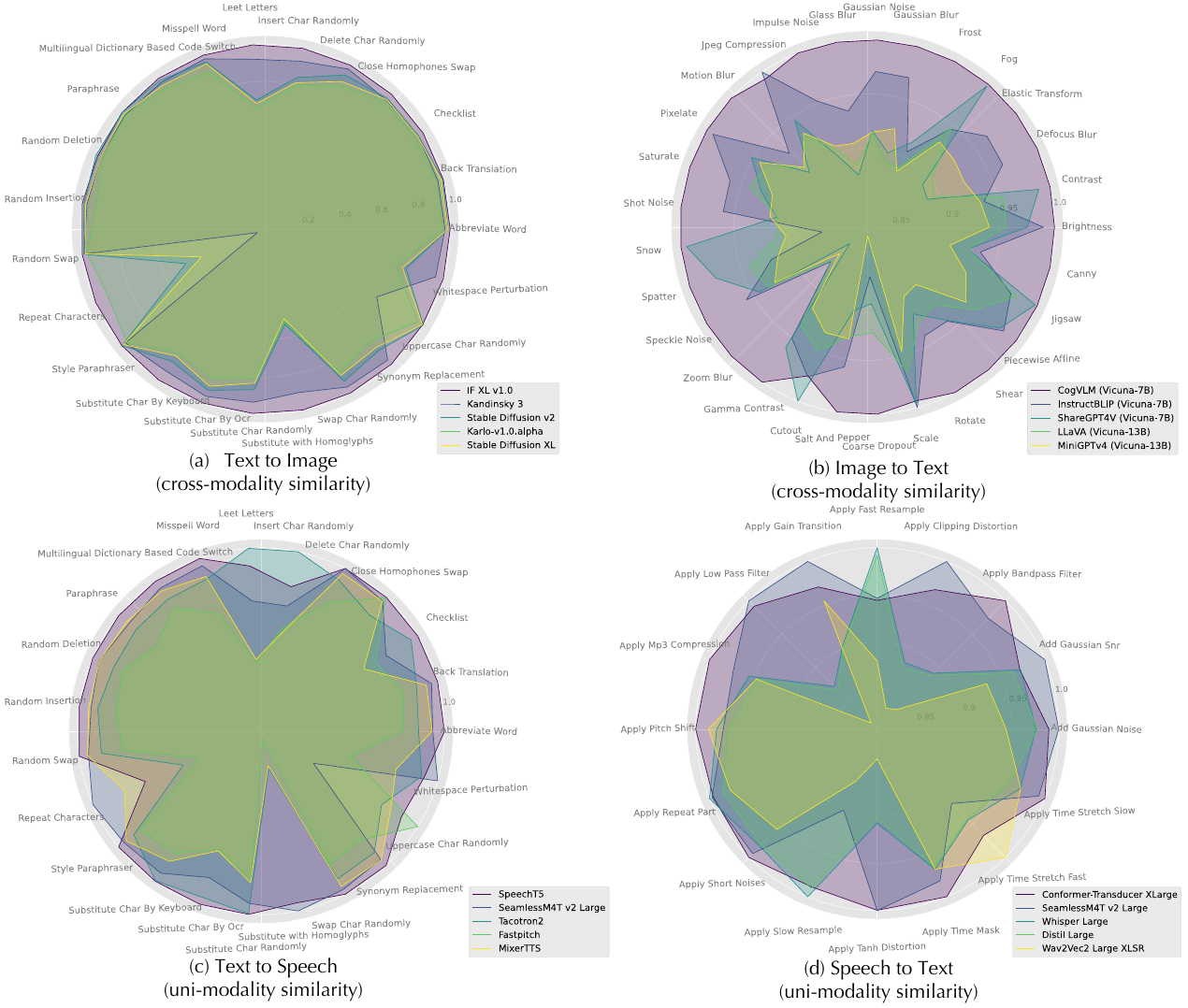}
  \vspace{-4mm}
\caption{Radar charts depicting the relative consistency scores of five selected models for various corruptions across four cross-modality tasks: (a) text-to-image, (b) image-to-text, (c) text-to-speech, and (d) speech-to-text. The scores are normalized with the highest scoring model set as the baseline (ratio of 1) for each type of corruption, allowing for a comparative analysis of each model's resilience to specific corruptions relative to the best performer in that category.}
  \label{fig:radar}
  \vspace{1mm}
\end{figure*}

\vspace{-2mm}
\section{Introduction}

The introduction of the CLIP framework~\cite{radford2021learning} has significantly accelerated recent advances in multimodal learning. The effective integration of visual and textual data by CLIP has not only marked a paradigm shift in machine learning's approach to diverse data modalities, but has also introduced unprecedented zero-shot prediction capabilities. Following it, models such as BLIP~\cite{li2022blip} and Stable Diffusion~\cite{rombach2022high} expanded on these ideas, improving the interaction between visual and textual elements and setting new standards for high-quality generative modeling.

The field has advanced significantly in recent years, with notable advances in large language models (LLMs)~\cite{brown2020language, openai2023gpt4}. Consequently, large multimodal models (LMMs) such as InstructBLIP~\cite{dai2023instructblip}, LLaVA~\cite{liu2023improved,liu2023visual}, MiniGPT-4~\cite{zhu2023minigpt}, CogVLM~\cite{wang2023cogvlm}, and ShareGPT4V~\cite{chen2023sharegpt4v}, building on the foundations of LLMs like LLaMA~\cite{touvron2023llama,touvron2023llama2} or Vicuna~\cite{vicuna2023,zheng2023judging}, have made remarkable progress in managing cross-modality interactions, particularly between image and text. This progress has resulted in the ability to perform more complex tasks, such as contextual understanding, multimodal conversation, and visual question answering, among others. With these advancements and a greater emphasis on practical applications, the robustness and real-world efficacy of these models are becoming increasingly important.

For early-stage multimodal models that do not involve LLMs, MMRobustness~\cite{qiu2022multimodal} evaluates the performance impact of common corruptions on the input modality across five downstream tasks. Following that, MME~\cite{fu2023mme} offers a comprehensive benchmark for evaluating LMMs, encompassing 14 subtasks in perception and cognition. Several efforts have also been made to investigate the possibility of manipulating LMMs via adversarial input images~\cite{bailey2023image,carlini2023aligned,dong2023robust,zhao2023evaluate}. Existing benchmarks, however, lack a comprehensive evaluation of the self-consistency of LMMs' outputs under common corruptions. Our goal is to bridge this gap by providing a comprehensive benchmark covering a wide range of popular LMMs.

To ensure efficient evaluation, we intend to exploit the diversity of large datasets such as LAION~\cite{schuhmann2021laion, schuhmann2022laion} and Common Voice~\cite{ardila2020common} while controlling the number of testing samples. To that end, we propose utilizing text as a semantic anchor. Because humans typically grasp complex concepts through text alone, this strategy is based on the belief that text adequately conveys semantic information across various modalities. We hope to select examples with significant textual changes for evaluation by mapping all modalities into text and calculating similarity as shown in the first row on~\Cref{fig:pipeline}.\looseness=-1

%\tianyu{Above done.}

% It then poses several critical questions, namely, \textit{how to select representative examples from large datasets for efficient evaluation} and \textit{what metric can be developed to consistently measure performance across different cross-modalities?}

% After selecting representative examples from the input modality, we apply a number of common corruptions to them. These corrupted inputs are then fed into LMMs to return outputs. The quantitative measurement of self-consistency performance in these models is a critical component of our benchmarking pipeline. One approach might be to measure self-consistency solely within the output modality, for example in an image-to-text task, calculating the similarity between captions generated from clean images versus those generated from corrupted images. However, if a model consistently generates poor-quality captions, the consistency score may be misleadingly high. To address this, we measure the relationship between the original, clean inputs and the outputs derived from the corrupted inputs to assess model self-consistency. In the image-to-text context, for example, we use the CLIP model to compare the similarity of the original clean images and the captions generated from the corresponding corrupted images. For a more complete reference, we also report the consistency score calculated solely on the output modality in Appendix~\ref{??}. 

After selecting representative examples from the input modality, we subject them to various common corruptions. These corrupted inputs are then fed into LMMs to generate outputs. A critical aspect of our evaluation is the quantitative measurement of self-consistency in these models. Primarily, we focus on cross-modality similarity—if such a model exists between the input and output modalities, it is utilized to measure self-consistency. However, in scenarios lacking a cross-modality model or the performance of such model is poor, we assess consistency within the output modality itself. This method, while straightforward, has its limitations; for instance, if a model consistently outputs poor-quality captions, it may still achieve a high consistency score on the output modality, which can be misleading. In our study, we present both cross-modality and output-only modality consistency scores for a comprehensive reference as shown in the second on~\Cref{fig:pipeline}, allowing for a nuanced interpretation of model performance.

%\jiawei{Since we put both results, the logic should change here, besides, its not always LMM... not sure if we should use it here, i prefer using Multimodal Models}

We mainly focus on cross-modality interactions involving text, image, and speech in this paper, encompassing four key generative tasks: \emph{text-to-image}, \emph{image-to-text}, \emph{text-to-speech}, and \emph{speech-to-text}. The released benchmark is named as \textbf{MMCBench} (\textbf{M}ulti\textbf{M}odal \textbf{C}orruption \textbf{Bench}mark). Our MMCBench performs extensive analyses of popular LMMs as well as traditional non-LLM-based multimodal models, providing insights into their robustness and self-consistency under various corruption scenarios. Specifically, we test against \textbf{23} text corruptions, \textbf{29} image corruptions, and \textbf{16} speech corruptions. For the text-to-image task, we assess \textbf{27} models across \textbf{37} different checkpoints; in the image-to-text category, \textbf{39} models are evaluated over \textbf{58} checkpoints; for text-to-speech, \textbf{14} models are scrutinized through \textbf{15} checkpoints; and for speech-to-text, we examine \textbf{41} models across \textbf{47} checkpoints. This extensive analysis offers deep insights into the resilience of these models in diverse and challenging conditions. 
% \jiawei{will add more models for text to speech in the future}

% The following are our contributions:

% \begin{itemize}
%     \item We develop \textbf{MMCBench} (\textbf{M}ulti\textbf{M}odal \textbf{C}orruption \textbf{Bench}mark), a comprehensive benchmark for measuring the self-consistency of LMMs against common corruptions;
%     \item We develop a methodology for selecting representative examples from large datasets, using text as a semantic anchor to evaluate the impact of common corruptions on multimodal model outputs;
%     \item We present a consistent metric system for assessing model performance across various cross-modalities, effectively quantifying these models' resilience to input corruptions;
%     \item Our evaluation includes an extensive analysis of several state-of-the-art multimodal models (both non-LLM and LLM-based), providing insights into their robustness and consistency under various corruption scenarios.
% \end{itemize}

\section{Related Work}
\label{relatedwork}

\textbf{Robustness evaluation on unimodal models.} 
Evaluating the robustness of unimodal models, such as vision and language models, is critical for their practical deployment.

Robustness is typically measured in the context of vision models along three key dimensions: \textit{common corruption}, \textit{adversarial robustness}, and \textit{distribution shift robustness}. Datasets such as ImageNet-C~\cite{hendrycks2018benchmarking} and 3DCC~\cite{kar20223d} are used to test whether classification models can maintain accurate predictions in the presence of common corruptions such as Gaussian noise, motion blur, and brightness changes. In terms of adversarial robustness, the emphasis shifts to how models respond to intentional perturbations. This can be evaluated using datasets like ImageNet-Patch~\cite{pintor2023imagenet}, which involves attaching malicious patches to images, or by subjecting models to direct $\ell_{p}$-norm attacks, as assessed by ARES~\cite{dong2020benchmarking} and RobustBench~\cite{croce2021robustbench}. The evaluation of distribution shift robustness tests the generalisation ability of vision models. Datasets like OOD-CV~\cite{zhao2022ood} and ImageNet-O~\cite{hendrycks2021natural} enable the evaluation of model performance on images that differ significantly from their training distribution. This evaluation helps assess the models' ability to adapt to new and unseen environments.

Similarly, in the context of language models, common corruption robustness can be evaluated by introducing, e.g., typos, word deletions, or whitespace insertions. On the other hand, adversarial robustness can be evaluated using model-based perturbations, as demonstrated by works such as BERT-Attack~\cite{li2020bert} and TextFooler~\cite{jin2020bert}. Platforms including TextAttack~\cite{morris2020textattack}, TextFlint~\cite{wang2021textflint}, Robustness Gym~\cite{goel2021robustness}, and NL-Augmenter~\cite{dhole2021nl} all provide convenient tools to facilitate these evaluations. The emergence of large language models (LLMs) has broadened the scope of robustness in language models. This expansion can be seen in platforms ranging from AdvGlue~\cite{wang2021adversarial}, which offers a comprehensive set of $14$ textual adversarial attacks at the word or sentence level, challenging language models in tasks like sentiment classification or natural language inference, to PromptBench~\cite{zhu2023promptbench}. PromptBench focuses on adversarial prompts that are intended mimic potential user errors, thereby testing the consistency of LLM outputs. There are also platforms like HADES~\cite{liu2021token} and HaluEval~\cite{li2023halueval}, which are aimed at detecting hallucinations in the contents generated by LLMs. Recently, DecodingTrust~\cite{wang2023decodingtrust} broadens its evaluation criteria to include toxicity, stereotype bias, privacy, machine ethics, and fairness.

\begin{table*}[!t]
\centering
\caption{Performance comparison of various \textbf{text-to-image} models evaluated by self-consistency scores (cross-modality) across different corruption intensities and data selection levels. Scores represent the average multiplied cosine similarities (max 2300) between original captions and the generated images for the captions under different corruption conditions.}
\vspace{0.2cm}
\label{tab:text2image_cross}
\begin{tabular*}{0.90\textwidth}{@{\extracolsep{\fill}}lcccccc}
\toprule
\multirow{2}{*}{Models} & \multicolumn{2}{c}{Hard} & \multicolumn{2}{c}{Random} & \multirow{2}{*}{Average} \\  \cmidrule(lr){2-3} \cmidrule(lr){4-5}
 & Heavy & Light & Heavy & Light & \\
\midrule
\first IF XL v1.0~\cite{deepfloyd_if_2023} & 630 & 722 & 577 & 679 & 652.00  \\
\second IF L v1.0~\cite{deepfloyd_if_2023} & 610 & 700 & 571 & 670 & 637.75 \\
\third IF M v1.0~\cite{deepfloyd_if_2023} & 584 & 666 & 564 & 661 & 618.75 \\
Kandinsky 3~\cite{arkhipkin2023kandinsky} & 563 & 664 & 555 & 670 & 613.00 \\
Stable Diffusion v2~\cite{rombach2022high} & 549 & 666 & 532 & 657 & 601.00 \\
SDXL Base~\cite{podell2023sdxl}    & 529 & 649 & 527 & 651 & 589.00 \\
SDXL Turbo~\cite{sauer2023adversarial} & 521 & 637 & 533 & 658 & 587.25 \\
SDXL Refiner~\cite{podell2023sdxl} & 526 & 645 & 522 & 645 & 584.50 \\
Karlo-v1.0.alpha~\cite{kakaobrain2022karlo-v1-alpha} & 520 & 627 & 522 & 637 & 576.50 \\
Openjourney v4~\cite{prompthero_openjourneyv4_2023} & 518 & 613 & 528 & 638 & 574.25 \\
LCM (SDXL)~\cite{luo2023latent, podell2023sdxl} & 515 & 626 & 515 & 638 & 573.50 \\
LCM LoRA (SDXL)~\cite{luo2023lcm, podell2023sdxl} & 503 & 613 & 517 & 640 & 568.25 \\
Stable Diffusion Turbo~\cite{sauer2023adversarial} & 493 & 602 & 519 & 641 & 563.75 \\
Anything Midjourney v4.1~\cite{prompthero_openjourneyv4_2023} & 491 & 595 & 510 & 623 & 554.75 \\
Dreamlike Photoreal 2.0~\cite{dreamlike_art_2023} & 491 & 590 & 513 & 623 & 554.25 \\
Stable Diffusion v1~\cite{rombach2022high} & 493 & 591 & 511 & 620 & 553.75 \\
Kandinsky 2.2~\cite{kandinsky2} & 488 & 588 & 510 & 627 & 553.25 \\
Small Stable Diffusion~\cite{kim2023architectural} & 473 & 565 & 518 & 625 & 545.25 \\
Unidiffuser~\cite{bao2023one} & 485 & 583 & 495 & 602 & 541.25 \\
LCM (SSD-1B)~\cite{luo2023latent, gupta2024progressive} & 468 & 564 & 490 & 599 & 530.25 \\
LCM LoRA (SSD-1B)~\cite{luo2023lcm, gupta2024progressive} & 462 & 560 & 485 & 600 & 526.75 \\
SSD 1B~\cite{gupta2024progressive} & 462 & 560 & 485 & 599 & 526.50 \\
LCM LoRA (SD v1)~\cite{luo2023lcm, rombach2022high} & 426  & 522  & 472  & 581  & 500.25  \\
Dreamshaper v7~\cite{lykon_dreamshaper7_2023} & 425 & 522 & 472 & 581 & 500.00 \\
LCM (Dreamshaper v7)~\cite{luo2023latent, lykon_dreamshaper7_2023} & 417 & 507 & 472 & 577 & 493.25 \\
Lafite~\cite{zhou2021lafite} & 415 & 480 & 474 & 558 & 481.75 \\
Glide~\cite{nichol2022glide} & 334 & 390 & 391 & 465 & 395.00\\
\bottomrule
\end{tabular*}
\vspace{6mm}
\end{table*}

\textbf{Robustness evaluation of multimodal models.}
A number of benchmarks for multimodal model robustness have been established, each focusing on a different aspect of model resilience~\cite{carlini2023aligned,zhao2023evaluate}. MMRobustness~\cite{qiu2022multimodal} primarily focuses on the relative performance drop observed when common corruptions are introduced to the input image or text. BenchLMM~\cite{cai2023benchlmm} instead investigates the robustness of large multimodal models (LMMs) such as GPT-4V~\cite{openai2023gpt4} and LLaVA~\cite{liu2023improved,liu2023visual}. It investigates how these models respond to stylistic shifts such as artistic, sensor, and application styles, thereby expanding our understanding of LMM resilience. The Bingo benchmark~\cite{cui2023holistic}, on the other hand, focuses on hallucination phenomena in visual language models. It delves into the biases and interference issues that are prevalent in LMMs, providing a specialized perspective on these unique challenges. The introduction of the VLAA framework~\cite{tu2023many}, which shifts the emphasis to out-of-distribution generalization and adversarial robustness, shedding light on critical vulnerabilities in LMMs. VLAA focuses on LMMs' responses to visually unrelated or linguistically perturbed inputs, as well as their resilience to adversarial attacks that threaten model safety and reliability. \looseness=-1

Despite these comprehensive benchmarks, a significant gap in the field remains: the lack of tools specifically designed to measure the consistency of multimodal model output when confronted with commonly corrupted inputs. Our research aims to close this gap, making contribution to the larger landscape of multimodal model evaluation.

% \begin{figure*}[th] 
%   \centering
%   \includegraphics[width=0.85\textwidth]{figures/text2image_example.pdf}
%   \caption{The generated images for the corrupted captions.}
%   \label{fig:text2image_example}
% \end{figure*}

\begin{table*}[!htbp]
\vspace{10mm}
\centering
\caption{Performance comparison of various \textbf{image-to-text} models evaluated by self-consistency scores (cross-modality) across different corruption intensities and data selection levels. Scores represent the average multiplied cosine similarities (max 2900) between original images and the captions generated for the images under different corruption conditions.}
\vspace{0.2cm}
\label{tab:image2text_cross}
\begin{threeparttable}
\begin{tabular*}{1.0\textwidth}{@{\extracolsep{\fill}}lccccccc}
\toprule
\multirow{2}{*}{Models} & \multirow{2}{*}{LLM} &\multicolumn{2}{c}{Hard} & \multicolumn{2}{c}{Random} & \multirow{2}{*}{Average} \\  \cmidrule(lr){3-4} \cmidrule(lr){5-6} 
&  & Heavy & Light & Heavy & Light & \\
\midrule
\first CogVLM~\cite{wang2023cogvlm,hong2023cogagent} & Vicuna 7B~\cite{zheng2023judging} & 815 & 887 & 841 & 886 & 857.25 \\
\second InstructBLIP~\cite{dai2023instructblip} & Vicuna 7B~\cite{zheng2023judging} & 760 & 856 & 799 & 848 & 815.75 \\
\third ShareGPT4V~\cite{chen2023sharegpt4v} & Vicuna 7B~\cite{zheng2023judging} & 724 & 843 & 798 & 853 & 804.50 \\
InstructBLIP~\cite{dai2023instructblip} & Flan T5 XXL~\cite{chung2022scaling}      & 751  & 841  & 774  & 820  & 796.50 \\
InstructBLIP~\cite{dai2023instructblip}  & Vicuna 13B~\cite{zheng2023judging} & 744 & 839 & 777 & 825 & 796.25 \\
LLaVA ~\cite{liu2023visual}  & Vicuna 13B~\cite{zheng2023judging} & 733 & 836 & 783 & 827 & 794.75 \\
LLaVA-v1.5~\cite{liu2023improved} & Vicuna 13B~\cite{zheng2023judging} & 736 & 827 & 788 & 827 & 794.50 \\
LLaVA-v1.5~\cite{liu2023improved} & Vicuna 7B~\cite{zheng2023judging} & 728 & 821 & 781 & 820 & 787.50 \\
MiniGPT-4~\cite{zhu2023minigpt} & Vicuna 13B~\cite{zheng2023judging} & 713 & 808 & 790 & 835 & 786.50 \\
BLIP2~\cite{li2023blip} & OPT-2.7b~\cite{zhang2022opt} & 704 & 790 & 791 & 840 & 781.25 \\
MiniGPT-4~\cite{zhu2023minigpt} & Vicuna 7B~\cite{zheng2023judging}  & 707 & 801 & 780 & 827 & 778.75 \\
BLIP2~\cite{li2023blip} &  OPT-6.7b~\cite{zhang2022opt}  & 703 & 792 & 784 & 833 & 778.00 \\
BLIP2~\cite{li2023blip} & Flan T5 XXL~\cite{chung2022scaling}  & 694 & 784 & 788 & 834 & 775.00 \\
BLIP2~\cite{li2023blip} & Flan T5 XL~\cite{chung2022scaling}   & 689 & 778 & 783 & 830 & 770.00 \\
LLaVA~\cite{liu2023improved} & LLaMA2 13B~\cite{touvron2023llama2}   & 700 & 803 & 763 & 810 & 769.00 \\
VisualGLM~\cite{du2022glm,ding2021cogview} & ChatGLM-6B~\cite{du2022glm, zeng2023glm-130b}  & 717 & 793 & 753 & 786 & 762.25 \\
MiniGPT-4~\cite{zhu2023minigpt} & LLaMA2 7B~\cite{touvron2023llama2}  & 685 & 767 & 773 & 810 & 758.75 \\
Qwen-VL-Chat~\cite{Qwen-VL}& Qwen-7B~\cite{bai2023qwen}  & 631 & 803 & 738 & 844 & 754.00 \\
LLaVA~\cite{liu2023improved} & LLaMA2 7B~\cite{touvron2023llama2}  & 687 & 786 & 747 & 796 & 754.00 \\
InstructBLIP~\cite{dai2023instructblip} & Flan T5 XL~\cite{chung2022scaling}         & 608  & 834  & 758  & 804  & 751.00 \\
LLaVA~\cite{liu2023improved} & MPT 7B~\cite{MosaicML2023Introducing}  & 681 & 778 & 744 & 791 & 748.50 \\
mPLUG-Owl2~\cite{ye2023mplug}  & LLaMA2 7B~\cite{touvron2023llama2}  & 608 & 746 & 722 & 806 & 720.50 \\
mPLUG-Owl~\cite{ye2023mplugv1}  & LLaMA 7B~\cite{touvron2023llama} & 642 & 742 & 723 & 770 & 719.25 \\
LLaMA-Adapter v2~\cite{gao2023llamaadapterv2}  & LLaMA 7B~\cite{touvron2023llama} & 646 & 744 & 712 & 759 & 715.25 \\
mPLUG-Owl (multilingual)~\cite{ye2023mplugv1}  & LLaMA 7B~\cite{touvron2023llama} & 607 & 700 & 693 & 741 & 685.25 \\
GIT Large~\cite{wang2022git} & -   & 581 & 683 & 696 & 753 & 678.25 \\
BLIP Large~\cite{li2022blip} & -  & 558 & 686 & 692 & 774 & 677.50 \\
OpenFlamingo~\cite{awadalla2023openflamingo,Alayrac2022FlamingoAV} & RedPajama 3B~\cite{redPajama3b}   & 600 & 685 & 681 & 731 & 674.25 \\
LaVIN~\cite{luo2023towards,luo2023cheap} & LLaMA 13B~\cite{touvron2023llama}  & 584 & 680 & 648 & 702 & 653.50 \\
Unidiffuser~\cite{bao2023one} & -  & 539 & 624 & 682 & 758 & 650.75 \\
OpenFlamingo~\cite{awadalla2023openflamingo,Alayrac2022FlamingoAV} & MPT 1B~\cite{mpt1b}  & 607 & 672 & 633 & 688 & 650.00 \\
BLIP Base~\cite{li2022blip} & -& 468 & 563 & 645 & 735 & 602.75 \\
GIT Base~\cite{wang2022git} & - & 468 & 546 & 630 & 698 & 585.50 \\
ImageBind-LLM~\cite{han2023imagebind} & Open Chinese LLaMA 7B~\cite{openlmlab, touvron2023llama}  & 489 & 569 & 613 & 654 & 581.25 \\
Multimodal-GPT~\cite{gong2023multimodalgpt} & LLaMA 7B~\cite{touvron2023llama}   & 513 & 573 & 547 & 575 & 552.00 \\
OpenFlamingo\tnote{1}~\cite{awadalla2023openflamingo,Alayrac2022FlamingoAV} & MPT 7B~\cite{MosaicML2023Introducing}  & 477 & 494 & 477 & 505 & 488.25 \\
ViT-GPT2~\cite{nlp_connect_2022} & GPT2~\cite{radford2019language}  & 357 & 399 & 503 & 551 & 452.50 \\
MiniGPT v2\tnote{2}~\cite{chen2023minigpt} & LLaMA2 7B~\cite{touvron2023llama2}   & 375 & 426 & 468 & 498 & 441.75 \\
% Pix2Struct Large~\cite{lee2023pix2struct} & -  & 190 & 220 & 180 & 218 & 202.00 \\
% Pix2Struct Base~\cite{lee2023pix2struct} & -   & 163 & 161 & 125 & 141 & 147.50 \\
\bottomrule
\end{tabular*}
\begin{tablenotes}
\footnotesize
\item[1] OpenFlamingo's inferior performance in instance tasks is attributed to the fact that their utilized language model, MPT 7B, is not tuned for instructions. Therefor, sometimes the model fails to follow our instructions accurately.
\item[2] MiniGPTv2's inferior performance in this task is due to the model's tendency to avoid providing captions for corrupted images.
\end{tablenotes}
\end{threeparttable}
\vspace{10mm}
\end{table*}

\section{Experiments}
% This section describes our approach to assessing output consistency in multimodal models. 
This section describes our MMCBench in details, including the common corruption methods used for each modality, the strategies of data selection, and the evaluation metrics (i.e., quantification of consistency scores) for each multimodal model.
% It includes our data selection strategy, the types of common corruption used in each modality, and the quantification of consistency scores for each multimodal model on the chosen data. 
In general, two primary modalities are considered in multimodal model evaluation: the input modality and the output modality. Our approach is twofold: first, we identify a representative subset from large base datasets such as LAION~\cite{laion-coco,schuhmann2022laion} for consistency evaluation; and second, we assess the consistency scores of multimodal models on the output modality when the input modality is corrupted.

\textbf{Data selection.} We use consistency scores in the text modality for ranking data samples for data selection, which is applicable to all cross-modalities, including image-to-text and speech-to-text. We calculate text similarities by mapping all data into the text modality and then rank and select samples with the lowest text similarity. This design choice is due to the critical role of text in representing semantic information across multiple modalities, as well as its utility in reducing computational complexity for selection. Furthermore, we include a comparative random sample from the base dataset. As a result, our data selection includes two levels: \emph{hard}: selecting the subset with the lowest text similarity post-corruption; and \emph{random}: randomly selecting a subset from the larger base dataset.

% In this work, we employ the \texttt{sentence-t5-large} model from sentence-transformer~\cite{reimers-2019-sentence-bert} to calculate text similarity.
\textbf{Model evaluation.} We apply various common corruptions to the input modality at different intensities (\emph{heavy} and \emph{light}) for the evaluation process. The consistency between the output modality's generated samples under input corruption and the original uncorrupted data is then measured. We generate four distinct result categories based on the dual levels of dataset and corruption intensity. Models are ranked according to their overall performance in these categories. Additionally, we will report the best performance of each model, particularly in cases where multiple checkpoints are available, such as those trained on different datasets.

\textbf{Disclaimer}: In this study, models were evaluated using greedy decoding wherever feasible, primarily to ensure reproducibility and to reduce computational costs. However, it is important to note that greedy decoding may not always yield the optimal output for each model, potentially leading to an underestimation of the true performance capabilities. Consequently, the results presented in our tables should be regarded as \textit{a conservative estimate or a lower bound} of the actual potential of the models.

Below, we introduce the common corruptions, data selection methods, and model evaluation criteria for each type of multimodal model in details.

\subsection{Text-to-Image Generation}
\label{sec:text-to-image}

\textbf{Text corruptions.} We incorporate a total of $23$ distinct types of text corruption, drawn from various platforms including NlpAug~\cite{ma2019nlpaug}, TextAugment~\cite{marivate2020improving}, and NL-Augmenter~\cite{dhole2021nl}. These corruptions are systematically categorized into three complexity levels, as follows:

\vspace{-1mm}
\begin{itemize}
    \item \textbf{Char Level:} \textit{Substitute Char by OCR}, \textit{Substitute Char by Keyboard}, \textit{Insert Char Randomly}, \textit{Substitute Char Randomly}, \textit{Swap Char Randomly}, \textit{Delete Char Randomly}, \textit{Uppercase Char Randomly}, \textit{Repeat Characters}, \textit{Leet Letters}~\cite{leet_letter}, \textit{Whitespace Perturbation}, \textit{Substitute with Homoglyphs}
    \vspace{-0mm}
    \item \textbf{Word Level:} \textit{Synonym Replacement}, \textit{Random Deletion}, \textit{Random Swap}, \textit{Random Insertion}, \textit{Misspell Word}, \textit{Abbreviate Word}, \textit{Multilingual Dictionary Based Code Switch}, \textit{Close Homophones Swap}
    \vspace{-0mm}
    \item \textbf{Sentence Level:} \textit{CheckList}~\cite{ribeiro-etal-2020-beyond}, \textit{Back Translation}~\cite{ng2019facebook,sugiyama-yoshinaga-2019-data}, \textit{Style Paraphraser}~\cite{style20}, \textit{Paraphrase}~\cite{chatgpt_paraphraser}
    \vspace{-0mm}
\end{itemize}
\vspace{-2mm}

Examples illustrating the corrupted text of each corruption type are provided in~\Cref{apx:tti_example}.

% \textbf{Base dataset.} 
% We begin by randomly selecting $10$ million caption-image pairs from the LAION-COCO dataset~\cite{laion-coco}, which originally contains $600$ million images. This dataset was chosen specifically for its large volume and high-quality COCO-style captions~\cite{lin2014microsoft}. Then, from these $10$ million pairs, we aim to narrow our focus to $1,000$ captions, chosen for their inherent complexity in describing the images' content. These captions are especially difficult for image generation models to handle, especially when it comes to maintaining consistency in the images generated after the captions have been corrupted in various ways.

% \textbf{Selection criteria.} Our goal is to select $1,000$ captions from the $10$ million dataset, focusing on those that present significant challenges in maintaining output consistency from caption to image under corruption. These captions are distinguished by their inherent complexity in language or syntax, as well as their richness in description. We create a selection criteria with four components to systematically evaluate each caption:

\textbf{Data selection.}
We start with a randomly selected subset of $10$ million caption-image pairs from the LAION-COCO dataset~\cite{laion-coco}, which originally contains $600$ million images. This dataset is chosen for its large volume and high-quality COCO-style captions~\cite{lin2014microsoft}. From these $10$ million pairs, we then select $1,000$ captions that present more challenges in maintaining output consistency in \emph{text-to-image} generation under various text corruptions. These captions are distinguished by their inherent complexity in language or syntax, as well as their richness in description. Specifically, we select them based on four scores:

1) \textit{The inconsistency score} is calculated by one minus the average cosine text similarity between the original captions and their heavily corrupted versions across the $23$ text corruption types. To compute text similarity, we utilize the \texttt{sentence-t5-large} model from sentence-transformer~\cite{reimers-2019-sentence-bert}. This metric quantifies the degree of semantic alteration in the captions due to corruption.

2) \textit{The readability score} is calculated using \texttt{textstat}\footnote{\scriptsize\url{https://github.com/textstat/textstat}} to evaluate the inherent semantic complexity of the captions.

3) \textit{The syntax complexity score} is measured by simply calculating the depth of the parse trees for each caption.

4) \textit{The description score} assesses the content's richness by counting \# of nouns, verbs, and adjectives in each caption.\looseness=-1

The latter three scores are normalized to a range of $0$ to $1$, ensuring a balanced evaluation of caption quality and complexity. The overall selection score for each caption is computed as a weighted sum of the four components, with the inconsistency score carrying a weight of $6$, and the other scores each having a weight of $1$. During the initial selection process, however, we discovered that captions with the highest selection scores typically describe diagrams, charts, or posters. To avoid selecting captions that are not semantically rich, we add an additional filtering step based on aesthetic scores. We first choose the top $10,000$ captions based on their selection scores, and then choose the $1,000$ captions with the highest aesthetics scores from this subset using the improved aesthetic predictor\footnote{\scriptsize\url{https://github.com/christophschuhmann/improved-aesthetic-predictor}}. This procedure ensures that the final caption selection is not only linguistically challenging, but also visually relevant and representative of the images. The image size is maintained at $256 \times 256$ pixels across all experiments.\looseness=-1

\textbf{Evaluation methodology.} Self-consistency of text-to-image models in the face of specified corruptions is quantitatively assessed using a defined metric: the average cosine similarity between the original captions and the images generated after applying the $23$ types of corruptions to the $1,000$ selected captions. This similarity score is then multiplied by $100$, resulting in a maximum possible value of $2,300$. To calculate this cross-modality similarity, we utilize the CLIP model\footnote{\scriptsize\url{https://huggingface.co/laion/CLIP-ViT-L-14-laion2B-s32B-b82K}} trained on the LAION-2B dataset. A higher score in this metric indicates that a model is more capable of dealing with common corruptions. The final evaluation results on various text-to-image models are shown in~\Cref{tab:text2image_cross}, while we also provide the evaluation results on uni-modality similarity in~\Cref{apx:tti_uni}.\looseness=-1

% \tianyu{Above done.}

As observed, the self-consistency scores for models processing our carefully selected captions (hard) tend to be lower than those for randomly sampled captions (random) across identical corruption levels. This trend validates the effectiveness of our selection criteria, which is based solely on text degradation without any interference from text-to-image models during the selection phase. Notably, high-performing models such as IF~\cite{deepfloyd_if_2023}, Kandinsky 3~\cite{arkhipkin2023kandinsky}, and Stable Diffusion v2~\cite{rombach2022high} exhibit a narrower performance gap between hard and random conditions, underscoring their robustness and superior handling of input corruptions.

\begin{table*}[!htbp]
\vspace{10mm}
\centering
\caption{Performance of various \textbf{speech-to-text} models evaluated by self-consistency scores (uni-modality) across different corruption intensities and data selection levels. Scores represent the average multiplied cosine similarities (max 1600) between the transcriptions generated based on the original speech and the transcriptions generated for the speeches under different corruption conditions.\looseness=-1}
\vspace{0.2cm}
\label{tab:speech2text_uni}
\begin{threeparttable}
\begin{tabular*}{0.85\textwidth}{@{\extracolsep{\fill}}lcccccc}
\toprule
\multirow{2}{*}{Models} & \multicolumn{2}{c}{Hard} & \multicolumn{2}{c}{Random} & \multirow{2}{*}{Average} \\  \cmidrule(lr){2-3} \cmidrule(lr){4-5}
 & Heavy & Light & Heavy & Light & \\
\midrule
\first Conformer-Transducer XLarge~\cite{gulati2020conformer}    & 1299 & 1488 & 1429 & 1551 & 1441.75 \\
\second SeamlessM4T v2-Large~\cite{barrault2023seamless} & 1281 & 1476 & 1420 & 1543 & 1430.00 \\
\third FastConformer-Transducer XXLarge~\cite{rekesh2023fast} & 1271 & 1477 & 1411 & 1549 & 1427.00 \\
Conformer-Transducer XXLarge~\cite{gulati2020conformer}   & 1274 & 1471 & 1414 & 1543 & 1425.50 \\
Conformer-CTC XLarge~\cite{gulati2020conformer} & 1272 & 1464 & 1407 & 1538 & 1420.25 \\
Conformer-Transducer Large~\cite{gulati2020conformer} & 1259 & 1457 & 1399 & 1537 & 1413.00 \\
SeamlessM4T Medium~\cite{seamlessm4t2023} & 1260 & 1450 & 1405 & 1531 & 1411.50 \\
FastConformer-CTC Large~\cite{rekesh2023fast} & 1250 & 1458 & 1394 & 1537 & 1409.75 \\
Conformer-CTC Large~\cite{gulati2020conformer} & 1251 & 1450 & 1394 & 1532 & 1406.75 \\
FastConformer-CTC XXLarge~\cite{rekesh2023fast} & 1245 & 1452 & 1391 & 1534 & 1405.50 \\
FastConformer-Transducer Large~\cite{rekesh2023fast}  & 1245 & 1450 & 1390 & 1534 & 1404.75 \\
FastConformer-CTC XLarge~\cite{rekesh2023fast} & 1237 & 1454 & 1384 & 1533 & 1402.00 \\
SeamlessM4T Large~\cite{seamlessm4t2023} & 1248 & 1442 & 1392 & 1525 & 1401.75 \\
FastConformer-Transducer XLarge~\cite{rekesh2023fast}  & 1228 & 1442 & 1376 & 1529 & 1393.75 \\
Whisper Large~\cite{radford2023robust} & 1211 & 1414 & 1395 & 1530 & 1387.50 \\
Distil Large~\cite{gandhi2023distilwhisper} & 1206 & 1402 & 1383 & 1521 & 1378.00 \\
Conformer-Transducer Medium~\cite{gulati2020conformer} & 1203 & 1399 & 1355 & 1504 & 1365.25 \\
Conformer-CTC Medium~\cite{gulati2020conformer} & 1203 & 1402 & 1353 & 1503 & 1365.25 \\
Whisper Medium~\cite{radford2023robust} & 1175 & 1384 & 1361 & 1519 & 1359.75 \\
Distil Medium~\cite{gandhi2023distilwhisper} & 1169 & 1376 & 1347 & 1509 & 1350.25 \\
Wav2Vec2 Large XLSR~\cite{conneau2020unsupervised, grosman2021xlsr53-large-english} & 1191 & 1383 & 1304 & 1489 & 1341.75 \\
Conformer-CTC Small~\cite{gulati2020conformer} & 1174 & 1361 & 1321 & 1477 & 1333.25 \\
Conformer-Transducer Small~\cite{gulati2020conformer} & 1168 & 1360 & 1316 & 1477 & 1330.25 \\
Whisper Small~\cite{radford2023robust} & 1141 & 1346 & 1323 & 1499 & 1327.25 \\
Distil Small~\cite{gandhi2023distilwhisper} & 1141 & 1347 & 1319 & 1494 & 1325.25 \\
MMS 1B All~\cite{pratap2023mms} & 1179 & 1355 & 1296 & 1467 & 1324.25 \\
Hubert Large~\cite{hsu2021hubert} & 1154 & 1335 & 1277 & 1465 & 1307.75 \\
Wav2Vec2 Conformer Rel Pos Large~\cite{wang2020fairseq} & 1142 & 1321 & 1265 & 1459 & 1296.75 \\
Wav2Vec2 Conformer Rope Large~\cite{wang2020fairseq} & 1142 & 1311 & 1261 & 1445 & 1289.75 \\
Robust Wav2Vec2 Large~\cite{hsu2021robust} & 1148 & 1311 & 1259 & 1440 & 1289.50 \\
Wav2Vec2 Large Self-Training~\cite{baevski2020wav2vec, xu2021self} & 1151 & 1298 & 1262 & 1432 & 1285.75 \\
S2T Small~\cite{wang2020fairseq} & 1182 & 1273 & 1250 & 1364 & 1267.25 \\
Whisper Base~\cite{radford2023robust} & 1084 & 1260 & 1263 & 1450 & 1264.25 \\
Wav2Vec2 Large~\cite{baevski2020wav2vec} & 1124 & 1244 & 1228 & 1390 & 1246.50 \\
Whisper Tiny~\cite{radford2023robust} & 1084 & 1219 & 1220 & 1404 & 1231.75 \\
Wav2Vec2 Base~\cite{baevski2020wav2vec} & 1118 & 1207 & 1212 & 1346 & 1220.75 \\
ESPnet~\cite{watanabe2018espnet, arora2023universlu, pmlr-v202-radford23a} & 1086 & 1232 & 1190 & 1364 & 1218.00 \\
S2T Medium~\cite{wang2020fairseq} & 1107 & 1211 & 1205 & 1328 & 1212.75 \\
SpeechT5~\cite{ao2022speecht5} & 1077 & 1195 & 1198 & 1359 & 1207.25 \\
S2T Large~\cite{wang2020fairseq} & 1087 & 1200 & 1188 & 1326 & 1200.25 \\
\bottomrule
\end{tabular*}
\begin{tablenotes}
\footnotesize
\item[1] The Conformer CTC/Transducers and FastConformer CTC/Transducers models here are sourced from NeMo~\cite{kuchaiev2019nemo}.
\end{tablenotes}
\end{threeparttable}
\vspace{10mm}
\end{table*}

\begin{table*}[!t]
\centering
\caption{Performance comparison of various \textbf{text-to-speech} models evaluated by self-consistency scores (uni-modality) across different corruption intensities and data selection levels. Scores represent the average multiplied cosine similarities (max 2300) between the transcriptions for the speeches generated for the clean transcriptions and the transcriptions for the speeches generated for the transcriptions under different corruption conditions. The name in parentheses indicates the source from which the checkpoint for this model was obtained.}
\vspace{0.2cm}
\label{tab:text2speech_uni}
\begin{tabular*}{0.925\textwidth}{@{\extracolsep{\fill}}lcccccc}
\toprule
\multirow{2}{*}{Models} & \multicolumn{2}{c}{Hard} & \multicolumn{2}{c}{Random} & \multirow{2}{*}{Average} \\  \cmidrule(lr){2-3} \cmidrule(lr){4-5}
 & Heavy & Light & Heavy & Light & \\
\midrule
\first SpeechT5~\cite{ao2022speecht5}            & 2023 & 2130 & 2007 & 2115 & 2068.75 \\
\second SeamlessM4T v2 Large~\cite{barrault2023seamless}      & 2007 & 2118 & 1989 & 2101 & 2053.75 \\
\third Tacotron2 (SpeechBrain)\tnote{1} \cite{shen2018natural, speechbrain}    & 2008 & 2111 & 1981 & 2083 & 2045.75 \\
Fastpitch (NeMo)~\cite{lancucki2021fastpitch, kuchaiev2019nemo}        & 1996 & 2103 & 1975 & 2083 & 2039.25 \\
MixerTTS (NeMo)~\cite{tatanov2022mixer, kuchaiev2019nemo}       & 1994 & 2101 & 1973 & 2082 & 2037.50 \\
Tacotron2 (NeMo)~\cite{shen2018natural, kuchaiev2019nemo}          & 1989 & 2098 & 1972 & 2080 & 2034.75 \\
MixerTTS-X (NeMo)\cite{tatanov2022mixer, kuchaiev2019nemo}    & 1985 & 2091 & 1965 & 2072 & 2028.25 \\
FastSpeech2 (Facebook)~\cite{ren2020fastspeech, wang2020fairseq}    & 1976 & 2076 & 1952 & 2053 & 2014.25 \\
XTTS v2~\cite{xtts}                & 1968 & 2068 & 1931 & 2027 & 1998.50 \\
SeamlessM4T Large~\cite{seamlessm4t2023}      & 1951 & 2056 & 1932 & 2043 & 1995.50 \\
ESPnet2~\cite{watanabe2018espnet} & 1963 & 2053 & 1928 & 2016 & 1990.00 \\
VITS (NeMo)~\cite{kim2021conditional, kuchaiev2019nemo}         & 1947 & 2043 & 1921 & 2014 & 1981.25 \\
MMS~\cite{pratap2023mms}                 & 1949 & 2039 & 1918 & 2007 & 1978.25 \\
SeamlessM4T Medium~\cite{seamlessm4t2023}      & 1875 & 1974 & 1880 & 1987 & 1929.00 \\
\bottomrule
\end{tabular*}
\vspace{5mm}
\end{table*}

\vspace{-2mm}
\subsection{Image-to-Text Generation}

\textbf{Image corruptions.}
In our robustness evaluation, we include a wide range of corruptions from the imagecorruptions library~\cite{michaelis2019dragon} which is based on ImageNet-C~\cite{hendrycks2018benchmarking}, and the imgaug library~\cite{imgaug}.  The selected types of corruptions are categorized as follows:
\vspace{-2mm}
\begin{itemize}
    \item \textbf{Noise-Related:} \textit{Gaussian Noise}, \textit{Shot Noise}, \textit{Impulse Noise}, \textit{Speckle Noise}
    \vspace{-1mm}
    \item \textbf{Blur-Related:} \textit{Defocus Blur}, \textit{Glass Blur}, \textit{Motion Blur}, \textit{Zoom Blur}, \textit{Gaussian Blur}
    \vspace{-1mm}
    \item \textbf{Weather Conditions:} \textit{Snow}, \textit{Frost}, \textit{Fog}
    \vspace{-1mm}
    \item \textbf{Digital:} \textit{Brightness}, \textit{Contrast}, \textit{Pixelate}, \textit{JPEG Compression}, \textit{Spatter}, \textit{Saturate}, \textit{Gamma Contrast}
    \vspace{-1mm}
    \item \textbf{Arithmetic:} \textit{Cutout}, \textit{Salt and Pepper}, \textit{Coarse Dropout}
    \vspace{-1mm}
    \item \textbf{Geometric:} \textit{Scale}, \textit{Rotate}, \textit{Shear}, \textit{Piecewise Affine}, \textit{Jigsaw}
    \vspace{-1mm}
    \item \textbf{Edge:} \textit{Canny}
    \vspace{-1mm}
\end{itemize}
\vspace{-2mm}
This integration results in a total of $29$ distinct and varied corruption methods for our comprehensive study. Examples illustrating the corrupted images of each corruption type are provided in~\Cref{apx:itt_example}.

% \textbf{Base dataset.} We choose a subset of $3$ million images from the LAION-Aesthetics dataset\footnote{\url{https://laion.ai/blog/laion-aesthetics/}} for our study. 
% % This selection is based on two criteria: high visual quality and the inherent challenge they present for multimodal models in generating correct captions under corrupt conditions.
% % We intend to select $1,000$ images from this subset, focusing on those that most significantly meet these criteria.

% \textbf{Selection criteria.} The selection score is based on the average cosine similarity between the text generated from the original uncorrupted images and the text generated from images subjected to ImageNet-C's $15$ common corruptions at severity level $3$. This assessment makes use of outputs from three baseline models: \texttt{vitgpt2}~\cite{nlp_connect_2022}, \texttt{blip-image-captioning-base}~\cite{li2022blip}, and \texttt{git-base}~\cite{wang2022git}. The $1,000$ images with the lowest text similarity are chosen.

\textbf{Data selection.} We select a subset of $3$ million images from the LAION-Aesthetics dataset\footnote{\scriptsize\url{https://laion.ai/blog/laion-aesthetics/}} for our study. We intend to select $1,000$ images from this subset, focusing on those with high visual quality and inherently pose challenges for multimodal models. The inconsistency score is determined by one minus the average cosine similarity between the text generated from the original uncorrupted images and the text generated from images subjected to the $15$ common corruptions of ImageNet-C with severity level 3. The $1,000$ images with the highest inconsistency score are chosen. This assessment utilizes outputs from three baseline models: \texttt{vitgpt2}~\cite{nlp_connect_2022}, \texttt{blip-base}~\cite{li2022blip}, and \texttt{git-base}~\cite{wang2022git}. The image size is maintained at $384 \times 384$ pixels across all experiments.

\textbf{Evaluation methodology.} Similarly, the self-consistency of image-to-text models in response to specified corruptions is quantified using a metric based on the average sum of the cosine similarities between the original images and the captions generated for the chosen $1,000$ images under the $29$ types of corruptions. This similarity score is then multiplied by $100$, resulting in a maximum possible value of $2,900$. For calculating this cross-modality similarity, we employ the same CLIP model as used in \Cref{sec:text-to-image}. A higher score indicates a model's superior ability to deal with common corruptions. Note that for MLLMs, we consistently employed the instruction \emph{``Describe this image as detailed as possible.''} Detailed evaluation results for various image-to-text models are presented in \Cref{tab:image2text_uni}, while the evaluation results on uni-modality similarity are deferred to~\Cref{apx:itt_uni}.\looseness=-1

The comparative analysis of model performances reveals that the self-consistency scores for models evaluated on our carefully selected images are significantly lower than those for models tested with randomly selected images, under the same levels of corruption, which attests to the effectiveness of our selection methodology. Besides, as we can see, large multimodal models (LMMs) consistently outperform their counterparts that do not utilize LLMs, highlighting the added resilience that LLM integration confers in the face of corrupted inputs. However, a larger LLM size does not necessarily equate to higher consistency; instead, some models with smaller LLMs (e.g., 7B model) actually achieve comparable or even superior scores than the LMMs equipped with larger LLMs (e.g., 13B model).

\vspace{-2mm}
\subsection{Text-to-Speech Generation}
\label{sec:text-to-speech}

\textbf{Text corruptions.} We use the same $23$ text corruptions as previously described in~\Cref{sec:text-to-image}, which is consistent with our text-to-image experiments.

% \textbf{Base dataset.} For this study, we use $1.75$ million validated text-speech pairs from the Common Voice 15.0 dataset~\cite{ardila2020common}. We meticulously selected a subset of $1,000$ pairs from this large collection, each chosen for its complexity and potential in challenging text-to-speech (TTS) models.

% \textbf{Text corruptions \& selection criteria.} We use the same $23$ text corruption types as previously described in~\Cref{sec:text-to-image}, which is consistent with our text-to-image experiments. Text-speech pairs are still chosen based on four scores: inconsistency, readability, syntax complexity, and description. This multifaceted approach ensures a thorough evaluation by focusing on pairs that pose significant linguistic and acoustic challenges to TTS models.

\textbf{Data selection.} For this study, we utilize $1.75$ million validated text-speech pairs from the Common Voice 15.0 dataset~\cite{ardila2020common}.
From this extensive collection, we extract a subset of $1,000$ pairs, each selected for its complexity and potential to challenge text-to-speech (TTS) models. 
Similar to text-to-image generation as described in~\Cref{sec:text-to-image}, the selection process is still based on the four scores: inconsistency, readability, syntax complexity, and description.
This multifaceted approach ensures a comprehensive evaluation by focusing on pairs that present significant linguistic and acoustic challenges to TTS models.

\textbf{Evaluation methodology.} When assessing the self-consistency of speech-to-text (STT) models, we initially considered cross-modality similarity using the CLAP model~\cite{elizalde2023clap} from LAION. However, it was determined to be less representative of true performance, given that CLAP's training predominantly involved classification-focused audio data, not speech. Consequently, we instead mainly focus to unimodality similarity here. This is measured as the average cosine similarity between the clean original transcriptions and the transcriptions of speech generated from texts with $16$ types of text corruptions. Transcription of the generated speech relies on the \texttt{wav2vec2-base-960h}~\cite{baevski2020wav2vec}, while the text similarity is still computed using the \texttt{sentence-t5-large} model from the sentence-transformer library. The resulting similarity scores are multiplied by $100$, yielding a maximum potential score of $2,300$. This method accounts for potential variations in speaker voices across STT models and places the emphasis on transcription consistency. The final evaluation results are shown in~\Cref{tab:text2speech_uni}; while for self-consistency scores on cross modality, we will include enhanced models for more accurate cross-similarity calculations between speech and text in the later version.

% while for the cross-modality self-consistency scores, which are not the main metric due to the outlined reasons, are still provided for completeness in \Cref{apx:tts_cross}.

%\tianyu{Above done.}
\vspace{-1mm}
The results shows that the level of corruption has a more pronounced impact on model performance compared to the selection method used for evaluation data. One potential explanation is that varying corruption levels in the input transcriptions may lead to differences in only certain segments of the generated speech. Consequently, the transcriptions of these speeches could differ by only a few characters or words. While the current models used for calculating text similarity are likely more attuned to semantic discrepancies rather than minor character or word changes, which could account for the observed differences in performance. We plan to propose more sensitive measures of similarity that can better capture these subtle variations in future work.

\vspace{-0mm}
\subsection{Speech-to-Text Generation}
\label{sec:speech-to-text}

\textbf{Speech corruptions.}
Our study incorporates $16$ distinct audio corruptions from audiomentations\footnote{\scriptsize\url{https://github.com/iver56/audiomentations}}, which are categorized as follows:
\vspace{-3mm}
\begin{itemize}
    \item \textbf{Noise Additions and Interference:} \textit{Gaussian Noise}, \textit{Short Noises}~\footnote{The short noise sounds are sampled from ESC-50
 dataset~\cite{piczak2015dataset}.}, \textit{Gaussian SNR}
    \item \textbf{Filtering and Frequency Adjustments:} \textit{Bandpass Filter}, \textit{Low Pass Filter}
    \item \textbf{Distortion and Audio Quality Effects:} \textit{Clipping Distortion}, \textit{MP3 Compression}, \textit{Tanh Distortion}
    \item \textbf{Temporal and Speed Modifications:} \textit{Fast Resample}, \textit{Slow Resample}, \textit{Time Stretch (Fast)}, \textit{Time Stretch (Slow)}
    \item \textbf{Pitch and Dynamic Range Adjustments:} \textit{Pitch Shift}, \textit{Gain Transition}
    \item \textbf{Repetitive and Temporal Effects:} \textit{Repeat Part}, \textit{Time Mask}
\end{itemize}
% \textbf{Base dataset.}
% We use the Common Voice 15.0 dataset~\cite{ardila2020common}, which includes $1.75$ million validated text-speech pairs and is known for its diversity. As a result, it is an ideal choice for speech-to-text (STT) analysis. We chose $1,000$ speeches from this large collection to demonstrate the difficulties in STT processing.

% \textbf{Selection criteria.}
% The selection process relies on the average cosine similarity between texts generated from corrupted and original audio, using baseline models such as \texttt{speecht5\_asr}~\cite{ao2022speecht5}, \texttt{wav2vec2-base-960h}~\cite{baevski2020wav2vec}, and \texttt{whisper-base.en}~\cite{radford2023robust}. We prioritize speeches with the greatest drop in text similarity after corruption.

\textbf{Data selection.} We use the Common Voice 15.0 dataset~\cite{ardila2020common}, which includes around $1.75$ million validated text-speech pairs and is known for its diversity. As a result, it is an ideal choice for speech-to-text (STT) analysis. We chose $1,000$ speeches from this large collection to demonstrate the difficulties in STT processing.
The selection process relies on the average cosine similarity between texts generated from corrupted and original audio, using baseline models such as \texttt{speecht5\_asr}~\cite{ao2022speecht5}, \texttt{wav2vec2-base-960h}~\cite{baevski2020wav2vec}, and \texttt{whisper-base.en}~\cite{radford2023robust}. We prioritize speeches with the greatest drop in text similarity after corruption. The original sampling rate for the audio is maintained at 16,000 Hz (16 kHz), which is a standard configuration for Automatic Speech Recognition (ASR) systems.

\vspace{-0mm}
\textbf{Evaluation methodology.}
For the speech-to-text (STT) models, our evaluation is still focused in uni-modality similarity, focusing on the consistency of the generated transcriptions amid speech corruptions. Specifically, we calculate the average cosine similarity between transcriptions generated from clean speech and those produced after applying $16$ distinct types of speech corruptions. These similarity scores are multiplied by $100$, allowing for a maximum score of $1,600$. A higher score suggests a model's increased robustness against speech corruption. The text similarity here is determined using the \texttt{sentence-t5-large} model from the sentence-transformer library. The corresponding results are presented in \Cref{tab:speech2text_uni}; while for self-consistency scores on cross modality, we will include enhanced models for more accurate cross-similarity calculations between speech and text in the later version.

% while we still provide the consistency score based on cross-modality in~\Cref{apx:stt_cross}.

%\tianyu{Above done.}
From the results, we can see that the scores for our selectively chosen speech samples are consistently lower than those for randomly sampled speech under the same levels of corruption, confirming the challenging nature of our selected samples across various TTS models.

\section{Discussion}
Returning self-consistent outputs in the face of input corruption is an essential requirement for the practical use of LMMs. We position MMCBench as a lightweight and comprehensive benchmark that concentrates on common corruptions in the multimodal learning literature. We intend to constantly update new models and incorporate more modalities, such as video, into MMCBench. Furthermore, future iterations of our research will aim to develop a more effective method for evaluating cross-modality similarity between speech and text. Regular updates and progress will be documented in our GitHub repository to facilitate community engagement and contributions.

%\newpage
\bibliography{ms}

\begin{thebibliography}{128}
\providecommand{\natexlab}[1]{#1}
\providecommand{\url}[1]{\texttt{#1}}
\expandafter\ifx\csname urlstyle\endcsname\relax
  \providecommand{\doi}[1]{doi: #1}\else
  \providecommand{\doi}{doi: \begingroup \urlstyle{rm}\Url}\fi

\bibitem[Alayrac et~al.(2022)Alayrac, Donahue, Luc, Miech, Barr, Hasson, Lenc, Mensch, Millican, Reynolds, Ring, Rutherford, Cabi, Han, Gong, Samangooei, Monteiro, Menick, Borgeaud, Brock, Nematzadeh, Sharifzadeh, Binkowski, Barreira, Vinyals, Zisserman, and Simonyan]{Alayrac2022FlamingoAV}
Alayrac, J.-B., Donahue, J., Luc, P., Miech, A., Barr, I., Hasson, Y., Lenc, K., Mensch, A., Millican, K., Reynolds, M., Ring, R., Rutherford, E., Cabi, S., Han, T., Gong, Z., Samangooei, S., Monteiro, M., Menick, J., Borgeaud, S., Brock, A., Nematzadeh, A., Sharifzadeh, S., Binkowski, M., Barreira, R., Vinyals, O., Zisserman, A., and Simonyan, K.
\newblock Flamingo: a visual language model for few-shot learning.
\newblock \emph{ArXiv}, abs/2204.14198, 2022.

\bibitem[Ao et~al.(2022)Ao, Wang, Zhou, Wang, Ren, Wu, Liu, Ko, Li, Zhang, et~al.]{ao2022speecht5}
Ao, J., Wang, R., Zhou, L., Wang, C., Ren, S., Wu, Y., Liu, S., Ko, T., Li, Q., Zhang, Y., et~al.
\newblock Speecht5: Unified-modal encoder-decoder pre-training for spoken language processing.
\newblock In \emph{Proceedings of the 60th Annual Meeting of the Association for Computational Linguistics (Volume 1: Long Papers)}, pp.\  5723--5738, 2022.

\bibitem[Ardila et~al.(2020)Ardila, Branson, Davis, Kohler, Meyer, Henretty, Morais, Saunders, Tyers, and Weber]{ardila2020common}
Ardila, R., Branson, M., Davis, K., Kohler, M., Meyer, J., Henretty, M., Morais, R., Saunders, L., Tyers, F., and Weber, G.
\newblock Common voice: A massively-multilingual speech corpus.
\newblock In \emph{Proceedings of the Twelfth Language Resources and Evaluation Conference}, pp.\  4218--4222, 2020.

\bibitem[Arkhipkin et~al.(2023)Arkhipkin, Filatov, Vasilev, Maltseva, Azizov, Pavlov, Agafonova, Kuznetsov, and Dimitrov]{arkhipkin2023kandinsky}
Arkhipkin, V., Filatov, A., Vasilev, V., Maltseva, A., Azizov, S., Pavlov, I., Agafonova, J., Kuznetsov, A., and Dimitrov, D.
\newblock Kandinsky 3.0 technical report, 2023.

\bibitem[Arora et~al.(2023)Arora, Futami, weon Jung, Peng, Sharma, Kashiwagi, Tsunoo, and Watanabe]{arora2023universlu}
Arora, S., Futami, H., weon Jung, J., Peng, Y., Sharma, R., Kashiwagi, Y., Tsunoo, E., and Watanabe, S.
\newblock Universlu: Universal spoken language understanding for diverse classification and sequence generation tasks with a single network, 2023.

\bibitem[Awadalla et~al.(2023)Awadalla, Gao, Gardner, Hessel, Hanafy, Zhu, Marathe, Bitton, Gadre, Sagawa, Jitsev, Kornblith, Koh, Ilharco, Wortsman, and Schmidt]{awadalla2023openflamingo}
Awadalla, A., Gao, I., Gardner, J., Hessel, J., Hanafy, Y., Zhu, W., Marathe, K., Bitton, Y., Gadre, S., Sagawa, S., Jitsev, J., Kornblith, S., Koh, P.~W., Ilharco, G., Wortsman, M., and Schmidt, L.
\newblock Openflamingo: An open-source framework for training large autoregressive vision-language models.
\newblock \emph{arXiv preprint arXiv:2308.01390}, 2023.

\bibitem[Baevski et~al.(2020)Baevski, Zhou, Mohamed, and Auli]{baevski2020wav2vec}
Baevski, A., Zhou, Y., Mohamed, A., and Auli, M.
\newblock wav2vec 2.0: A framework for self-supervised learning of speech representations.
\newblock \emph{Advances in neural information processing systems}, 33:\penalty0 12449--12460, 2020.

\bibitem[Bai et~al.(2023{\natexlab{a}})Bai, Bai, Chu, Cui, Dang, Deng, Fan, Ge, Han, Huang, et~al.]{bai2023qwen}
Bai, J., Bai, S., Chu, Y., Cui, Z., Dang, K., Deng, X., Fan, Y., Ge, W., Han, Y., Huang, F., et~al.
\newblock Qwen technical report.
\newblock \emph{arXiv preprint arXiv:2309.16609}, 2023{\natexlab{a}}.

\bibitem[Bai et~al.(2023{\natexlab{b}})Bai, Bai, Yang, Wang, Tan, Wang, Lin, Zhou, and Zhou]{Qwen-VL}
Bai, J., Bai, S., Yang, S., Wang, S., Tan, S., Wang, P., Lin, J., Zhou, C., and Zhou, J.
\newblock Qwen-vl: A frontier large vision-language model with versatile abilities.
\newblock \emph{arXiv preprint arXiv:2308.12966}, 2023{\natexlab{b}}.

\bibitem[Bailey et~al.(2023)Bailey, Ong, Russell, and Emmons]{bailey2023image}
Bailey, L., Ong, E., Russell, S., and Emmons, S.
\newblock Image hijacks: Adversarial images can control generative models at runtime.
\newblock \emph{arXiv preprint arXiv:2309.00236}, 2023.

\bibitem[Bao et~al.(2023)Bao, Nie, Xue, Li, Pu, Wang, Yue, Cao, Su, and Zhu]{bao2023one}
Bao, F., Nie, S., Xue, K., Li, C., Pu, S., Wang, Y., Yue, G., Cao, Y., Su, H., and Zhu, J.
\newblock One transformer fits all distributions in multi-modal diffusion at scale.
\newblock \emph{arXiv preprint arXiv:2303.06555}, 2023.

\bibitem[Barrault et~al.(2023{\natexlab{a}})Barrault, Chung, Meglioli, Dale, Dong, Duppenthaler, Duquenne, Ellis, Elsahar, Haaheim, et~al.]{barrault2023seamless}
Barrault, L., Chung, Y.-A., Meglioli, M.~C., Dale, D., Dong, N., Duppenthaler, M., Duquenne, P.-A., Ellis, B., Elsahar, H., Haaheim, J., et~al.
\newblock Seamless: Multilingual expressive and streaming speech translation.
\newblock \emph{arXiv preprint arXiv:2312.05187}, 2023{\natexlab{a}}.

\bibitem[Barrault et~al.(2023{\natexlab{b}})Barrault, Chung, Meglioli, Dale, Dong, Duquenne, Elsahar, Gong, Heffernan, Hoffman, et~al.]{seamlessm4t2023}
Barrault, L., Chung, Y.-A., Meglioli, M.~C., Dale, D., Dong, N., Duquenne, P.-A., Elsahar, H., Gong, H., Heffernan, K., Hoffman, J., et~al.
\newblock Seamlessm4t-massively multilingual \& multimodal machine translation.
\newblock \emph{arXiv preprint arXiv:2308.11596}, 2023{\natexlab{b}}.

\bibitem[Brown et~al.(2020)Brown, Mann, Ryder, Subbiah, Kaplan, Dhariwal, Neelakantan, Shyam, Sastry, Askell, et~al.]{brown2020language}
Brown, T., Mann, B., Ryder, N., Subbiah, M., Kaplan, J.~D., Dhariwal, P., Neelakantan, A., Shyam, P., Sastry, G., Askell, A., et~al.
\newblock Language models are few-shot learners.
\newblock \emph{Advances in neural information processing systems}, 33:\penalty0 1877--1901, 2020.

\bibitem[Cai et~al.(2023)Cai, Song, Guan, Chen, Luo, Yi, and Kot]{cai2023benchlmm}
Cai, R., Song, Z., Guan, D., Chen, Z., Luo, X., Yi, C., and Kot, A.
\newblock Benchlmm: Benchmarking cross-style visual capability of large multimodal models.
\newblock \emph{arXiv preprint arXiv:2312.02896}, 2023.

\bibitem[Carlini et~al.(2023)Carlini, Nasr, Choquette-Choo, Jagielski, Gao, Awadalla, Koh, Ippolito, Lee, Tramer, et~al.]{carlini2023aligned}
Carlini, N., Nasr, M., Choquette-Choo, C.~A., Jagielski, M., Gao, I., Awadalla, A., Koh, P.~W., Ippolito, D., Lee, K., Tramer, F., et~al.
\newblock Are aligned neural networks adversarially aligned?
\newblock \emph{arXiv preprint arXiv:2306.15447}, 2023.

\bibitem[Chen et~al.(2023{\natexlab{a}})Chen, Zhu, Shen, Li, Liu, Zhang, Krishnamoorthi, Chandra, Xiong, and Elhoseiny]{chen2023minigpt}
Chen, J., Zhu, D., Shen, X., Li, X., Liu, Z., Zhang, P., Krishnamoorthi, R., Chandra, V., Xiong, Y., and Elhoseiny, M.
\newblock Minigpt-v2: large language model as a unified interface for vision-language multi-task learning.
\newblock \emph{arXiv preprint arXiv:2310.09478}, 2023{\natexlab{a}}.

\bibitem[Chen et~al.(2023{\natexlab{b}})Chen, Li, Dong, Zhang, He, Wang, Zhao, and Lin]{chen2023sharegpt4v}
Chen, L., Li, J., Dong, X., Zhang, P., He, C., Wang, J., Zhao, F., and Lin, D.
\newblock Sharegpt4v: Improving large multi-modal models with better captions.
\newblock \emph{arXiv preprint arXiv:2311.12793}, 2023{\natexlab{b}}.

\bibitem[Chiang et~al.(2023)Chiang, Li, Lin, Sheng, Wu, Zhang, Zheng, Zhuang, Zhuang, Gonzalez, Stoica, and Xing]{vicuna2023}
Chiang, W.-L., Li, Z., Lin, Z., Sheng, Y., Wu, Z., Zhang, H., Zheng, L., Zhuang, S., Zhuang, Y., Gonzalez, J.~E., Stoica, I., and Xing, E.~P.
\newblock Vicuna: An open-source chatbot impressing gpt-4 with 90\%* chatgpt quality, March 2023.
\newblock URL \url{https://lmsys.org/blog/2023-03-30-vicuna/}.

\bibitem[Chung et~al.(2022)Chung, Hou, Longpre, Zoph, Tay, Fedus, Li, Wang, Dehghani, Brahma, et~al.]{chung2022scaling}
Chung, H.~W., Hou, L., Longpre, S., Zoph, B., Tay, Y., Fedus, W., Li, Y., Wang, X., Dehghani, M., Brahma, S., et~al.
\newblock Scaling instruction-finetuned language models.
\newblock \emph{arXiv preprint arXiv:2210.11416}, 2022.

\bibitem[Conneau et~al.(2020)Conneau, Baevski, Collobert, Mohamed, and Auli]{conneau2020unsupervised}
Conneau, A., Baevski, A., Collobert, R., Mohamed, A., and Auli, M.
\newblock Unsupervised cross-lingual representation learning for speech recognition.
\newblock \emph{arXiv preprint arXiv:2006.13979}, 2020.

\bibitem[coqui(2023)]{xtts}
coqui.
\newblock Xtts-v2.
\newblock \url{https://huggingface.co/coqui/XTTS-v2/tree/main}, 2023.

\bibitem[Croce et~al.(2021)Croce, Andriushchenko, Sehwag, Debenedetti, Flammarion, Chiang, Mittal, and Hein]{croce2021robustbench}
Croce, F., Andriushchenko, M., Sehwag, V., Debenedetti, E., Flammarion, N., Chiang, M., Mittal, P., and Hein, M.
\newblock Robustbench: a standardized adversarial robustness benchmark.
\newblock In \emph{Thirty-fifth Conference on Neural Information Processing Systems Datasets and Benchmarks Track (Round 2)}, 2021.

\bibitem[Cui et~al.(2023)Cui, Zhou, Yang, Wu, Zhang, Zou, and Yao]{cui2023holistic}
Cui, C., Zhou, Y., Yang, X., Wu, S., Zhang, L., Zou, J., and Yao, H.
\newblock Holistic analysis of hallucination in gpt-4v (ision): Bias and interference challenges.
\newblock \emph{arXiv preprint arXiv:2311.03287}, 2023.

\bibitem[Dai et~al.(2023)Dai, Li, Li, Tiong, Zhao, Wang, Li, Fung, and Hoi]{dai2023instructblip}
Dai, W., Li, J., Li, D., Tiong, A. M.~H., Zhao, J., Wang, W., Li, B., Fung, P., and Hoi, S.
\newblock Instructblip: Towards general-purpose vision-language models with instruction tuning, 2023.

\bibitem[deep floyd(2023)]{deepfloyd_if_2023}
deep floyd.
\newblock If.
\newblock \url{https://github.com/deep-floyd/IF}, 2023.

\bibitem[Dhole et~al.(2021)Dhole, Gangal, Gehrmann, Gupta, Li, Mahamood, Mahendiran, Mille, Shrivastava, Tan, et~al.]{dhole2021nl}
Dhole, K.~D., Gangal, V., Gehrmann, S., Gupta, A., Li, Z., Mahamood, S., Mahendiran, A., Mille, S., Shrivastava, A., Tan, S., et~al.
\newblock Nl-augmenter: A framework for task-sensitive natural language augmentation.
\newblock \emph{arXiv preprint arXiv:2112.02721}, 2021.

\bibitem[Ding et~al.(2021)Ding, Yang, Hong, Zheng, Zhou, Yin, Lin, Zou, Shao, Yang, et~al.]{ding2021cogview}
Ding, M., Yang, Z., Hong, W., Zheng, W., Zhou, C., Yin, D., Lin, J., Zou, X., Shao, Z., Yang, H., et~al.
\newblock Cogview: Mastering text-to-image generation via transformers.
\newblock In \emph{Advances in Neural Information Processing Systems (NeurIPS)}, 2021.

\bibitem[Dong et~al.(2020)Dong, Fu, Yang, Pang, Su, Xiao, and Zhu]{dong2020benchmarking}
Dong, Y., Fu, Q.-A., Yang, X., Pang, T., Su, H., Xiao, Z., and Zhu, J.
\newblock Benchmarking adversarial robustness on image classification.
\newblock In \emph{IEEE Conference on Computer Vision and Pattern Recognition (CVPR)}, 2020.

\bibitem[Dong et~al.(2023)Dong, Chen, Chen, Fang, Yang, Zhang, Tian, Su, and Zhu]{dong2023robust}
Dong, Y., Chen, H., Chen, J., Fang, Z., Yang, X., Zhang, Y., Tian, Y., Su, H., and Zhu, J.
\newblock How robust is google's bard to adversarial image attacks?
\newblock \emph{arXiv preprint arXiv:2309.11751}, 2023.

\bibitem[{Dreamlike Art}(2023)]{dreamlike_art_2023}
{Dreamlike Art}.
\newblock Dreamlike photoreal 2.0.
\newblock \url{https://huggingface.co/dreamlike-art/dreamlike-photoreal-2.0}, 2023.

\bibitem[Du et~al.(2022)Du, Qian, Liu, Ding, Qiu, Yang, and Tang]{du2022glm}
Du, Z., Qian, Y., Liu, X., Ding, M., Qiu, J., Yang, Z., and Tang, J.
\newblock Glm: General language model pretraining with autoregressive blank infilling.
\newblock In \emph{Proceedings of the 60th Annual Meeting of the Association for Computational Linguistics (Volume 1: Long Papers)}, pp.\  320--335, 2022.

\bibitem[Eger et~al.(2019)Eger, Sahin, R{\"{u}}ckl{\'{e}}, Lee, Schulz, Mesgar, Swarnkar, Simpson, and Gurevych]{leet_letter}
Eger, S., Sahin, G.~G., R{\"{u}}ckl{\'{e}}, A., Lee, J., Schulz, C., Mesgar, M., Swarnkar, K., Simpson, E., and Gurevych, I.
\newblock Text processing like humans do: Visually attacking and shielding {NLP} systems.
\newblock \emph{CoRR}, abs/1903.11508, 2019.
\newblock URL \url{http://arxiv.org/abs/1903.11508}.

\bibitem[Elizalde et~al.(2023)Elizalde, Deshmukh, Al~Ismail, and Wang]{elizalde2023clap}
Elizalde, B., Deshmukh, S., Al~Ismail, M., and Wang, H.
\newblock Clap learning audio concepts from natural language supervision.
\newblock In \emph{ICASSP 2023-2023 IEEE International Conference on Acoustics, Speech and Signal Processing (ICASSP)}, pp.\  1--5. IEEE, 2023.

\bibitem[Fu et~al.(2023)Fu, Chen, Shen, Qin, Zhang, Lin, Yang, Zheng, Li, Sun, et~al.]{fu2023mme}
Fu, C., Chen, P., Shen, Y., Qin, Y., Zhang, M., Lin, X., Yang, J., Zheng, X., Li, K., Sun, X., et~al.
\newblock Mme: A comprehensive evaluation benchmark for multimodal large language models.
\newblock \emph{arXiv preprint arXiv:2306.13394}, 2023.

\bibitem[Gandhi et~al.(2023)Gandhi, von Platen, and Rush]{gandhi2023distilwhisper}
Gandhi, S., von Platen, P., and Rush, A.~M.
\newblock Distil-whisper: Robust knowledge distillation via large-scale pseudo labelling, 2023.

\bibitem[Gao et~al.(2023)Gao, Han, Zhang, Lin, Geng, Zhou, Zhang, Lu, He, Yue, Li, and Qiao]{gao2023llamaadapterv2}
Gao, P., Han, J., Zhang, R., Lin, Z., Geng, S., Zhou, A., Zhang, W., Lu, P., He, C., Yue, X., Li, H., and Qiao, Y.
\newblock Llama-adapter v2: Parameter-efficient visual instruction model.
\newblock \emph{arXiv preprint arXiv:2304.15010}, 2023.

\bibitem[Goel et~al.(2021)Goel, Rajani, Vig, Taschdjian, Bansal, and R{\'e}]{goel2021robustness}
Goel, K., Rajani, N.~F., Vig, J., Taschdjian, Z., Bansal, M., and R{\'e}, C.
\newblock Robustness gym: Unifying the nlp evaluation landscape.
\newblock In \emph{Proceedings of the 2021 Conference of the North American Chapter of the Association for Computational Linguistics: Human Language Technologies: Demonstrations}, pp.\  42--55, 2021.

\bibitem[Gong et~al.(2023)Gong, Lyu, Zhang, Wang, Zheng, Zhao, Liu, Zhang, Luo, and Chen]{gong2023multimodalgpt}
Gong, T., Lyu, C., Zhang, S., Wang, Y., Zheng, M., Zhao, Q., Liu, K., Zhang, W., Luo, P., and Chen, K.
\newblock Multimodal-gpt: A vision and language model for dialogue with humans, 2023.

\bibitem[Grosman(2021)]{grosman2021xlsr53-large-english}
Grosman, J.
\newblock Fine-tuned {XLSR}-53 large model for speech recognition in {E}nglish.
\newblock \url{https://huggingface.co/jonatasgrosman/wav2vec2-large-xlsr-53-english}, 2021.

\bibitem[Gulati et~al.(2020)Gulati, Qin, Chiu, Parmar, Zhang, Yu, Han, Wang, Zhang, Wu, et~al.]{gulati2020conformer}
Gulati, A., Qin, J., Chiu, C.-C., Parmar, N., Zhang, Y., Yu, J., Han, W., Wang, S., Zhang, Z., Wu, Y., et~al.
\newblock Conformer: Convolution-augmented transformer for speech recognition.
\newblock \emph{Interspeech 2020}, 2020.

\bibitem[Gupta et~al.(2024)Gupta, Jaddipal, Prabhala, Paul, and Platen]{gupta2024progressive}
Gupta, Y., Jaddipal, V.~V., Prabhala, H., Paul, S., and Platen, P.~V.
\newblock Progressive knowledge distillation of stable diffusion xl using layer level loss, 2024.

\bibitem[Han et~al.(2023)Han, Zhang, Shao, Gao, Xu, Xiao, Zhang, Liu, Wen, Guo, et~al.]{han2023imagebind}
Han, J., Zhang, R., Shao, W., Gao, P., Xu, P., Xiao, H., Zhang, K., Liu, C., Wen, S., Guo, Z., et~al.
\newblock Imagebind-llm: Multi-modality instruction tuning.
\newblock \emph{arXiv preprint arXiv:2309.03905}, 2023.

\bibitem[Hendrycks \& Dietterich(2018)Hendrycks and Dietterich]{hendrycks2018benchmarking}
Hendrycks, D. and Dietterich, T.
\newblock Benchmarking neural network robustness to common corruptions and perturbations.
\newblock In \emph{International Conference on Learning Representations (ICLR)}, 2018.

\bibitem[Hendrycks et~al.(2021)Hendrycks, Zhao, Basart, Steinhardt, and Song]{hendrycks2021natural}
Hendrycks, D., Zhao, K., Basart, S., Steinhardt, J., and Song, D.
\newblock Natural adversarial examples.
\newblock In \emph{Proceedings of the IEEE/CVF Conference on Computer Vision and Pattern Recognition}, pp.\  15262--15271, 2021.

\bibitem[Hong et~al.(2023)Hong, Wang, Lv, Xu, Yu, Ji, Wang, Wang, Dong, Ding, and Tang]{hong2023cogagent}
Hong, W., Wang, W., Lv, Q., Xu, J., Yu, W., Ji, J., Wang, Y., Wang, Z., Dong, Y., Ding, M., and Tang, J.
\newblock Cogagent: A visual language model for gui agents, 2023.

\bibitem[Hsu et~al.(2021{\natexlab{a}})Hsu, Bolte, Tsai, Lakhotia, Salakhutdinov, and Mohamed]{hsu2021hubert}
Hsu, W.-N., Bolte, B., Tsai, Y.-H.~H., Lakhotia, K., Salakhutdinov, R., and Mohamed, A.
\newblock Hubert: Self-supervised speech representation learning by masked prediction of hidden units.
\newblock \emph{IEEE/ACM Transactions on Audio, Speech, and Language Processing}, 29:\penalty0 3451--3460, 2021{\natexlab{a}}.

\bibitem[Hsu et~al.(2021{\natexlab{b}})Hsu, Sriram, Baevski, Likhomanenko, Xu, Pratap, Kahn, Lee, Collobert, Synnaeve, et~al.]{hsu2021robust}
Hsu, W.-N., Sriram, A., Baevski, A., Likhomanenko, T., Xu, Q., Pratap, V., Kahn, J., Lee, A., Collobert, R., Synnaeve, G., et~al.
\newblock Robust wav2vec 2.0: Analyzing domain shift in self-supervised pre-training.
\newblock \emph{arXiv preprint arXiv:2104.01027}, 2021{\natexlab{b}}.

\bibitem[Jin et~al.(2020)Jin, Jin, Zhou, and Szolovits]{jin2020bert}
Jin, D., Jin, Z., Zhou, J.~T., and Szolovits, P.
\newblock Is bert really robust? a strong baseline for natural language attack on text classification and entailment.
\newblock In \emph{Proceedings of the AAAI conference on artificial intelligence}, 2020.

\bibitem[Jung et~al.(2020)Jung, Wada, Crall, Tanaka, Graving, Reinders, Yadav, Banerjee, Vecsei, Kraft, Rui, Borovec, Vallentin, Zhydenko, Pfeiffer, Cook, Fernández, De~Rainville, Weng, Ayala-Acevedo, Meudec, Laporte, et~al.]{imgaug}
Jung, A.~B., Wada, K., Crall, J., Tanaka, S., Graving, J., Reinders, C., Yadav, S., Banerjee, J., Vecsei, G., Kraft, A., Rui, Z., Borovec, J., Vallentin, C., Zhydenko, S., Pfeiffer, K., Cook, B., Fernández, I., De~Rainville, F.-M., Weng, C.-H., Ayala-Acevedo, A., Meudec, R., Laporte, M., et~al.
\newblock {imgaug}.
\newblock \url{https://github.com/aleju/imgaug}, 2020.
\newblock Online; accessed 01-Feb-2020.

\bibitem[Kar et~al.(2022)Kar, Yeo, Atanov, and Zamir]{kar20223d}
Kar, O.~F., Yeo, T., Atanov, A., and Zamir, A.
\newblock 3d common corruptions and data augmentation.
\newblock In \emph{Proceedings of the IEEE/CVF Conference on Computer Vision and Pattern Recognition}, pp.\  18963--18974, 2022.

\bibitem[Kim et~al.(2023)Kim, Song, Castells, and Choi]{kim2023architectural}
Kim, B.-K., Song, H.-K., Castells, T., and Choi, S.
\newblock On architectural compression of text-to-image diffusion models.
\newblock \emph{arXiv preprint arXiv:2305.15798}, 2023.

\bibitem[Kim et~al.(2021)Kim, Kong, and Son]{kim2021conditional}
Kim, J., Kong, J., and Son, J.
\newblock Conditional variational autoencoder with adversarial learning for end-to-end text-to-speech.
\newblock In \emph{International Conference on Machine Learning}, pp.\  5530--5540. PMLR, 2021.

\bibitem[Krishna et~al.(2020)Krishna, Wieting, and Iyyer]{style20}
Krishna, K., Wieting, J., and Iyyer, M.
\newblock Reformulating unsupervised style transfer as paraphrase generation.
\newblock In \emph{Empirical Methods in Natural Language Processing}, 2020.

\bibitem[Kuchaiev et~al.(2019)Kuchaiev, Li, Nguyen, Hrinchuk, Leary, Ginsburg, Kriman, Beliaev, Lavrukhin, Cook, et~al.]{kuchaiev2019nemo}
Kuchaiev, O., Li, J., Nguyen, H., Hrinchuk, O., Leary, R., Ginsburg, B., Kriman, S., Beliaev, S., Lavrukhin, V., Cook, J., et~al.
\newblock Nemo: a toolkit for building ai applications using neural modules.
\newblock \emph{arXiv preprint arXiv:1909.09577}, 2019.

\bibitem[Laion(2022)]{laion-coco}
Laion.
\newblock {Laion-COCO}: A large-scale custom object dataset for computer vision, 2022.
\newblock URL \url{https://laion.ai/blog/laion-coco/}.

\bibitem[{\L}a{\'n}cucki(2021)]{lancucki2021fastpitch}
{\L}a{\'n}cucki, A.
\newblock Fastpitch: Parallel text-to-speech with pitch prediction.
\newblock In \emph{ICASSP 2021-2021 IEEE International Conference on Acoustics, Speech and Signal Processing (ICASSP)}, pp.\  6588--6592. IEEE, 2021.

\bibitem[Lee et~al.(2022)Lee, Kim, Choi, Kim, Byeon, Baek, and Kim]{kakaobrain2022karlo-v1-alpha}
Lee, D., Kim, J., Choi, J., Kim, J., Byeon, M., Baek, W., and Kim, S.
\newblock Karlo-v1.0.alpha on coyo-100m and cc15m.
\newblock \url{https://github.com/kakaobrain/karlo}, 2022.

\bibitem[Li et~al.(2022)Li, Li, Xiong, and Hoi]{li2022blip}
Li, J., Li, D., Xiong, C., and Hoi, S.
\newblock Blip: Bootstrapping language-image pre-training for unified vision-language understanding and generation.
\newblock In \emph{International Conference on Machine Learning}, pp.\  12888--12900. PMLR, 2022.

\bibitem[Li et~al.(2023{\natexlab{a}})Li, Cheng, Zhao, Nie, and Wen]{li2023halueval}
Li, J., Cheng, X., Zhao, W.~X., Nie, J.-Y., and Wen, J.-R.
\newblock Halueval: A large-scale hallucination evaluation benchmark for large language models.
\newblock In \emph{Proceedings of the 2023 Conference on Empirical Methods in Natural Language Processing}, pp.\  6449--6464, 2023{\natexlab{a}}.

\bibitem[Li et~al.(2023{\natexlab{b}})Li, Li, Savarese, and Hoi]{li2023blip}
Li, J., Li, D., Savarese, S., and Hoi, S.
\newblock Blip-2: Bootstrapping language-image pre-training with frozen image encoders and large language models.
\newblock \emph{arXiv preprint arXiv:2301.12597}, 2023{\natexlab{b}}.

\bibitem[Li et~al.(2020)Li, Ma, Guo, Xue, and Qiu]{li2020bert}
Li, L., Ma, R., Guo, Q., Xue, X., and Qiu, X.
\newblock Bert-attack: Adversarial attack against bert using bert.
\newblock In \emph{Proceedings of the 2020 Conference on Empirical Methods in Natural Language Processing (EMNLP)}, pp.\  6193--6202, 2020.

\bibitem[Lin et~al.(2014)Lin, Maire, Belongie, Hays, Perona, Ramanan, Doll{\'a}r, and Zitnick]{lin2014microsoft}
Lin, T.-Y., Maire, M., Belongie, S., Hays, J., Perona, P., Ramanan, D., Doll{\'a}r, P., and Zitnick, C.~L.
\newblock Microsoft coco: Common objects in context.
\newblock In \emph{Computer Vision--ECCV 2014: 13th European Conference, Zurich, Switzerland, September 6-12, 2014, Proceedings, Part V 13}, pp.\  740--755. Springer, 2014.

\bibitem[Liu et~al.(2023{\natexlab{a}})Liu, Li, Li, and Lee]{liu2023improved}
Liu, H., Li, C., Li, Y., and Lee, Y.~J.
\newblock Improved baselines with visual instruction tuning.
\newblock \emph{arXiv preprint arXiv:2310.03744}, 2023{\natexlab{a}}.

\bibitem[Liu et~al.(2023{\natexlab{b}})Liu, Li, Wu, and Lee]{liu2023visual}
Liu, H., Li, C., Wu, Q., and Lee, Y.~J.
\newblock Visual instruction tuning.
\newblock \emph{arXiv preprint arXiv:2304.08485}, 2023{\natexlab{b}}.

\bibitem[Liu et~al.(2021)Liu, Zhang, Brockett, Mao, Sui, Chen, and Dolan]{liu2021token}
Liu, T., Zhang, Y., Brockett, C., Mao, Y., Sui, Z., Chen, W., and Dolan, B.
\newblock A token-level reference-free hallucination detection benchmark for free-form text generation.
\newblock \emph{arXiv preprint arXiv:2104.08704}, 2021.

\bibitem[Luo et~al.(2023{\natexlab{a}})Luo, Huang, Zhou, Sun, Jiang, Wang, and Ji]{luo2023towards}
Luo, G., Huang, M., Zhou, Y., Sun, X., Jiang, G., Wang, Z., and Ji, R.
\newblock Towards efficient visual adaption via structural re-parameterization.
\newblock \emph{arXiv preprint arXiv:2302.08106}, 2023{\natexlab{a}}.

\bibitem[Luo et~al.(2023{\natexlab{b}})Luo, Zhou, Ren, Chen, Sun, and Ji]{luo2023cheap}
Luo, G., Zhou, Y., Ren, T., Chen, S., Sun, X., and Ji, R.
\newblock Cheap and quick: Efficient vision-language instruction tuning for large language models.
\newblock \emph{arXiv preprint arXiv:2305.15023}, 2023{\natexlab{b}}.

\bibitem[Luo et~al.(2023{\natexlab{c}})Luo, Tan, Huang, Li, and Zhao]{luo2023latent}
Luo, S., Tan, Y., Huang, L., Li, J., and Zhao, H.
\newblock Latent consistency models: Synthesizing high-resolution images with few-step inference, 2023{\natexlab{c}}.

\bibitem[Luo et~al.(2023{\natexlab{d}})Luo, Tan, Patil, Gu, von Platen, Passos, Huang, Li, and Zhao]{luo2023lcm}
Luo, S., Tan, Y., Patil, S., Gu, D., von Platen, P., Passos, A., Huang, L., Li, J., and Zhao, H.
\newblock Lcm-lora: A universal stable-diffusion acceleration module.
\newblock \emph{arXiv preprint arXiv:2311.05556}, 2023{\natexlab{d}}.

\bibitem[Lykon(2023)]{lykon_dreamshaper7_2023}
Lykon.
\newblock dreamshaper-7.
\newblock \url{https://huggingface.co/Lykon/dreamshaper-7}, 2023.

\bibitem[Ma(2019)]{ma2019nlpaug}
Ma, E.
\newblock Nlp augmentation.
\newblock https://github.com/makcedward/nlpaug, 2019.

\bibitem[Marivate \& Sefara(2020)Marivate and Sefara]{marivate2020improving}
Marivate, V. and Sefara, T.
\newblock Improving short text classification through global augmentation methods.
\newblock In \emph{International Cross-Domain Conference for Machine Learning and Knowledge Extraction}, pp.\  385--399. Springer, 2020.

\bibitem[Michaelis et~al.(2019)Michaelis, Mitzkus, Geirhos, Rusak, Bringmann, Ecker, Bethge, and Brendel]{michaelis2019dragon}
Michaelis, C., Mitzkus, B., Geirhos, R., Rusak, E., Bringmann, O., Ecker, A.~S., Bethge, M., and Brendel, W.
\newblock Benchmarking robustness in object detection: Autonomous driving when winter is coming.
\newblock \emph{arXiv preprint arXiv:1907.07484}, 2019.

\bibitem[Morris et~al.(2020)Morris, Lifland, Yoo, Grigsby, Jin, and Qi]{morris2020textattack}
Morris, J.~X., Lifland, E., Yoo, J.~Y., Grigsby, J., Jin, D., and Qi, Y.
\newblock Textattack: A framework for adversarial attacks, data augmentation, and adversarial training in nlp.
\newblock \emph{arXiv preprint arXiv:2005.05909}, 2020.

\bibitem[mosaicml(2023)]{mpt1b}
mosaicml.
\newblock mpt-1b-redpajama-200b.
\newblock \url{https://huggingface.co/mosaicml/mpt-1b-redpajama-200b}, 2023.

\bibitem[Ng et~al.(2019)Ng, Yee, Baevski, Ott, Auli, and Edunov]{ng2019facebook}
Ng, N., Yee, K., Baevski, A., Ott, M., Auli, M., and Edunov, S.
\newblock Facebook fair's wmt19 news translation task submission.
\newblock \emph{arXiv preprint arXiv:1907.06616}, 2019.

\bibitem[Nichol et~al.(2022)Nichol, Dhariwal, Ramesh, Shyam, Mishkin, Mcgrew, Sutskever, and Chen]{nichol2022glide}
Nichol, A.~Q., Dhariwal, P., Ramesh, A., Shyam, P., Mishkin, P., Mcgrew, B., Sutskever, I., and Chen, M.
\newblock Glide: Towards photorealistic image generation and editing with text-guided diffusion models.
\newblock In \emph{International Conference on Machine Learning}, pp.\  16784--16804. PMLR, 2022.

\bibitem[{NLP Connect}(2022)]{nlp_connect_2022}
{NLP Connect}.
\newblock vit-gpt2-image-captioning (revision 0e334c7), 2022.
\newblock URL \url{https://huggingface.co/nlpconnect/vit-gpt2-image-captioning}.

\bibitem[OpenAI(2023)]{openai2023gpt4}
OpenAI.
\newblock Gpt-4 technical report, 2023.

\bibitem[openlmlab(2023)]{openlmlab}
openlmlab.
\newblock open-chinese-llama-7b-patch.
\newblock \url{https://huggingface.co/openlmlab/open-chinese-llama-7b-patch}, 2023.

\bibitem[Piczak(2015)]{piczak2015dataset}
Piczak, K.~J.
\newblock Esc: Dataset for environmental sound classification.
\newblock In \emph{Proceedings of the 23rd ACM international conference on Multimedia}, pp.\  1015--1018, 2015.

\bibitem[Pintor et~al.(2023)Pintor, Angioni, Sotgiu, Demetrio, Demontis, Biggio, and Roli]{pintor2023imagenet}
Pintor, M., Angioni, D., Sotgiu, A., Demetrio, L., Demontis, A., Biggio, B., and Roli, F.
\newblock Imagenet-patch: A dataset for benchmarking machine learning robustness against adversarial patches.
\newblock \emph{Pattern Recognition}, 134:\penalty0 109064, 2023.

\bibitem[Podell et~al.(2023)Podell, English, Lacey, Blattmann, Dockhorn, M{\"u}ller, Penna, and Rombach]{podell2023sdxl}
Podell, D., English, Z., Lacey, K., Blattmann, A., Dockhorn, T., M{\"u}ller, J., Penna, J., and Rombach, R.
\newblock Sdxl: Improving latent diffusion models for high-resolution image synthesis.
\newblock \emph{arXiv preprint arXiv:2307.01952}, 2023.

\bibitem[Pratap et~al.(2023)Pratap, Tjandra, Shi, Tomasello, Babu, Kundu, Elkahky, Ni, Vyas, Fazel-Zarandi, Baevski, Adi, Zhang, Hsu, Conneau, and Auli]{pratap2023mms}
Pratap, V., Tjandra, A., Shi, B., Tomasello, P., Babu, A., Kundu, S., Elkahky, A., Ni, Z., Vyas, A., Fazel-Zarandi, M., Baevski, A., Adi, Y., Zhang, X., Hsu, W.-N., Conneau, A., and Auli, M.
\newblock Scaling speech technology to 1,000+ languages.
\newblock \emph{arXiv}, 2023.

\bibitem[prompthero(2023)]{prompthero_openjourneyv4_2023}
prompthero.
\newblock openjourney-v4.
\newblock \url{https://huggingface.co/prompthero/openjourney-v4}, 2023.

\bibitem[Qiu et~al.(2022)Qiu, Zhu, Shi, Wenzel, Tang, Zhao, Li, and Li]{qiu2022multimodal}
Qiu, J., Zhu, Y., Shi, X., Wenzel, F., Tang, Z., Zhao, D., Li, B., and Li, M.
\newblock Are multimodal models robust to image and text perturbations?
\newblock \emph{arXiv preprint arXiv:2212.08044}, 2022.

\bibitem[Radford et~al.(2019)Radford, Wu, Child, Luan, Amodei, Sutskever, et~al.]{radford2019language}
Radford, A., Wu, J., Child, R., Luan, D., Amodei, D., Sutskever, I., et~al.
\newblock Language models are unsupervised multitask learners.
\newblock \emph{OpenAI blog}, 1\penalty0 (8):\penalty0 9, 2019.

\bibitem[Radford et~al.(2021)Radford, Kim, Hallacy, Ramesh, Goh, Agarwal, Sastry, Askell, Mishkin, Clark, et~al.]{radford2021learning}
Radford, A., Kim, J.~W., Hallacy, C., Ramesh, A., Goh, G., Agarwal, S., Sastry, G., Askell, A., Mishkin, P., Clark, J., et~al.
\newblock Learning transferable visual models from natural language supervision.
\newblock In \emph{International conference on machine learning}, pp.\  8748--8763. PMLR, 2021.

\bibitem[Radford et~al.(2023{\natexlab{a}})Radford, Kim, Xu, Brockman, Mcleavey, and Sutskever]{pmlr-v202-radford23a}
Radford, A., Kim, J.~W., Xu, T., Brockman, G., Mcleavey, C., and Sutskever, I.
\newblock Robust speech recognition via large-scale weak supervision.
\newblock In Krause, A., Brunskill, E., Cho, K., Engelhardt, B., Sabato, S., and Scarlett, J. (eds.), \emph{Proceedings of the 40th International Conference on Machine Learning}, volume 202 of \emph{Proceedings of Machine Learning Research}, pp.\  28492--28518. PMLR, 23--29 Jul 2023{\natexlab{a}}.

\bibitem[Radford et~al.(2023{\natexlab{b}})Radford, Kim, Xu, Brockman, McLeavey, and Sutskever]{radford2023robust}
Radford, A., Kim, J.~W., Xu, T., Brockman, G., McLeavey, C., and Sutskever, I.
\newblock Robust speech recognition via large-scale weak supervision.
\newblock In \emph{International Conference on Machine Learning}, pp.\  28492--28518. PMLR, 2023{\natexlab{b}}.

\bibitem[Ravanelli et~al.(2021)Ravanelli, Parcollet, Plantinga, Rouhe, Cornell, Lugosch, Subakan, Dawalatabad, Heba, Zhong, Chou, Yeh, Fu, Liao, Rastorgueva, Grondin, Aris, Na, Gao, Mori, and Bengio]{speechbrain}
Ravanelli, M., Parcollet, T., Plantinga, P., Rouhe, A., Cornell, S., Lugosch, L., Subakan, C., Dawalatabad, N., Heba, A., Zhong, J., Chou, J.-C., Yeh, S.-L., Fu, S.-W., Liao, C.-F., Rastorgueva, E., Grondin, F., Aris, W., Na, H., Gao, Y., Mori, R.~D., and Bengio, Y.
\newblock {SpeechBrain}: A general-purpose speech toolkit, 2021.
\newblock arXiv:2106.04624.

\bibitem[Reimers \& Gurevych(2019)Reimers and Gurevych]{reimers-2019-sentence-bert}
Reimers, N. and Gurevych, I.
\newblock Sentence-bert: Sentence embeddings using siamese bert-networks.
\newblock In \emph{Proceedings of the 2019 Conference on Empirical Methods in Natural Language Processing}. Association for Computational Linguistics, 11 2019.
\newblock URL \url{https://arxiv.org/abs/1908.10084}.

\bibitem[Rekesh et~al.(2023)Rekesh, Kriman, Majumdar, Noroozi, Juang, Hrinchuk, Kumar, and Ginsburg]{rekesh2023fast}
Rekesh, D., Kriman, S., Majumdar, S., Noroozi, V., Juang, H., Hrinchuk, O., Kumar, A., and Ginsburg, B.
\newblock Fast conformer with linearly scalable attention for efficient speech recognition.
\newblock \emph{arXiv preprint arXiv:2305.05084}, 2023.

\bibitem[Ren et~al.(2020)Ren, Hu, Tan, Qin, Zhao, Zhao, and Liu]{ren2020fastspeech}
Ren, Y., Hu, C., Tan, X., Qin, T., Zhao, S., Zhao, Z., and Liu, T.-Y.
\newblock Fastspeech 2: Fast and high-quality end-to-end text to speech.
\newblock In \emph{International Conference on Learning Representations}, 2020.

\bibitem[Ribeiro et~al.(2020)Ribeiro, Wu, Guestrin, and Singh]{ribeiro-etal-2020-beyond}
Ribeiro, M.~T., Wu, T., Guestrin, C., and Singh, S.
\newblock Beyond accuracy: Behavioral testing of {NLP} models with {C}heck{L}ist.
\newblock In Jurafsky, D., Chai, J., Schluter, N., and Tetreault, J. (eds.), \emph{Proceedings of the 58th Annual Meeting of the Association for Computational Linguistics}, pp.\  4902--4912, Online, July 2020. Association for Computational Linguistics.
\newblock \doi{10.18653/v1/2020.acl-main.442}.
\newblock URL \url{https://aclanthology.org/2020.acl-main.442}.

\bibitem[Rombach et~al.(2022)Rombach, Blattmann, Lorenz, Esser, and Ommer]{rombach2022high}
Rombach, R., Blattmann, A., Lorenz, D., Esser, P., and Ommer, B.
\newblock High-resolution image synthesis with latent diffusion models.
\newblock In \emph{Proceedings of the IEEE/CVF conference on computer vision and pattern recognition}, pp.\  10684--10695, 2022.

\bibitem[Sauer et~al.(2023)Sauer, Lorenz, Blattmann, and Rombach]{sauer2023adversarial}
Sauer, A., Lorenz, D., Blattmann, A., and Rombach, R.
\newblock Adversarial diffusion distillation.
\newblock \emph{arXiv preprint arXiv:2311.17042}, 2023.

\bibitem[Schuhmann et~al.(2021)Schuhmann, Vencu, Beaumont, Kaczmarczyk, Mullis, Katta, Coombes, Jitsev, and Komatsuzaki]{schuhmann2021laion}
Schuhmann, C., Vencu, R., Beaumont, R., Kaczmarczyk, R., Mullis, C., Katta, A., Coombes, T., Jitsev, J., and Komatsuzaki, A.
\newblock Laion-400m: Open dataset of clip-filtered 400 million image-text pairs.
\newblock \emph{arXiv preprint arXiv:2111.02114}, 2021.

\bibitem[Schuhmann et~al.(2022)Schuhmann, Beaumont, Vencu, Gordon, Wightman, Cherti, Coombes, Katta, Mullis, Wortsman, et~al.]{schuhmann2022laion}
Schuhmann, C., Beaumont, R., Vencu, R., Gordon, C., Wightman, R., Cherti, M., Coombes, T., Katta, A., Mullis, C., Wortsman, M., et~al.
\newblock Laion-5b: An open large-scale dataset for training next generation image-text models.
\newblock In \emph{Advances in Neural Information Processing Systems (NeurIPS)}, 2022.

\bibitem[Shakhmatov et~al.(2023)Shakhmatov, Razzhigaev, Nikolich, Arkhipkin, Pavlov, Kuznetsov, and Dimitrov]{kandinsky2}
Shakhmatov, A., Razzhigaev, A., Nikolich, A., Arkhipkin, V., Pavlov, I., Kuznetsov, A., and Dimitrov, D.
\newblock kandinsky 2.2.
\newblock \url{https://huggingface.co/kandinsky-community/kandinsky-2-2-decoder}, 2023.

\bibitem[Shen et~al.(2018)Shen, Pang, Weiss, Schuster, Jaitly, Yang, Chen, Zhang, Wang, Skerrv-Ryan, et~al.]{shen2018natural}
Shen, J., Pang, R., Weiss, R.~J., Schuster, M., Jaitly, N., Yang, Z., Chen, Z., Zhang, Y., Wang, Y., Skerrv-Ryan, R., et~al.
\newblock Natural tts synthesis by conditioning wavenet on mel spectrogram predictions.
\newblock In \emph{2018 IEEE international conference on acoustics, speech and signal processing (ICASSP)}, pp.\  4779--4783. IEEE, 2018.

\bibitem[Sugiyama \& Yoshinaga(2019)Sugiyama and Yoshinaga]{sugiyama-yoshinaga-2019-data}
Sugiyama, A. and Yoshinaga, N.
\newblock Data augmentation using back-translation for context-aware neural machine translation.
\newblock In Popescu-Belis, A., Lo{\'a}iciga, S., Hardmeier, C., and Xiong, D. (eds.), \emph{Proceedings of the Fourth Workshop on Discourse in Machine Translation (DiscoMT 2019)}, pp.\  35--44, Hong Kong, China, November 2019. Association for Computational Linguistics.
\newblock \doi{10.18653/v1/D19-6504}.
\newblock URL \url{https://aclanthology.org/D19-6504}.

\bibitem[Tatanov et~al.(2022)Tatanov, Beliaev, and Ginsburg]{tatanov2022mixer}
Tatanov, O., Beliaev, S., and Ginsburg, B.
\newblock Mixer-tts: non-autoregressive, fast and compact text-to-speech model conditioned on language model embeddings.
\newblock In \emph{ICASSP 2022-2022 IEEE International Conference on Acoustics, Speech and Signal Processing (ICASSP)}, pp.\  7482--7486. IEEE, 2022.

\bibitem[Team et~al.(2023)]{MosaicML2023Introducing}
Team, M. et~al.
\newblock Introducing mpt-7b: a new standard for open-source, commercially usable llms, 2023.

\bibitem[togethercomputer(2023)]{redPajama3b}
togethercomputer.
\newblock Redpajama-incite-instruct-3b-v1.
\newblock \url{https://huggingface.co/togethercomputer/RedPajama-INCITE-Instruct-3B-v1}, 2023.

\bibitem[Touvron et~al.(2023{\natexlab{a}})Touvron, Lavril, Izacard, Martinet, Lachaux, Lacroix, Rozi{\`e}re, Goyal, Hambro, Azhar, et~al.]{touvron2023llama}
Touvron, H., Lavril, T., Izacard, G., Martinet, X., Lachaux, M.-A., Lacroix, T., Rozi{\`e}re, B., Goyal, N., Hambro, E., Azhar, F., et~al.
\newblock Llama: Open and efficient foundation language models.
\newblock \emph{arXiv preprint arXiv:2302.13971}, 2023{\natexlab{a}}.

\bibitem[Touvron et~al.(2023{\natexlab{b}})Touvron, Martin, Stone, Albert, Almahairi, Babaei, Bashlykov, Batra, Bhargava, Bhosale, et~al.]{touvron2023llama2}
Touvron, H., Martin, L., Stone, K., Albert, P., Almahairi, A., Babaei, Y., Bashlykov, N., Batra, S., Bhargava, P., Bhosale, S., et~al.
\newblock Llama 2: Open foundation and fine-tuned chat models.
\newblock \emph{arXiv preprint arXiv:2307.09288}, 2023{\natexlab{b}}.

\bibitem[Tu et~al.(2023)Tu, Cui, Wang, Zhou, Zhao, Han, Zhou, Yao, and Xie]{tu2023many}
Tu, H., Cui, C., Wang, Z., Zhou, Y., Zhao, B., Han, J., Zhou, W., Yao, H., and Xie, C.
\newblock How many unicorns are in this image? a safety evaluation benchmark for vision llms.
\newblock \emph{arXiv preprint arXiv:2311.16101}, 2023.

\bibitem[Vorobev \& Kuznetsov(2023)Vorobev and Kuznetsov]{chatgpt_paraphraser}
Vorobev, M. K.~V. and Kuznetsov, M.
\newblock A paraphrasing model based on chatgpt paraphrases.
\newblock \emph{A paraphrasing model based on ChatGPT paraphrases}, 2023.

\bibitem[Wang et~al.(2021{\natexlab{a}})Wang, Xu, Wang, Gan, Cheng, Gao, Awadallah, and Li]{wang2021adversarial}
Wang, B., Xu, C., Wang, S., Gan, Z., Cheng, Y., Gao, J., Awadallah, A.~H., and Li, B.
\newblock Adversarial glue: A multi-task benchmark for robustness evaluation of language models.
\newblock In \emph{Thirty-fifth Conference on Neural Information Processing Systems Datasets and Benchmarks Track (Round 2)}, 2021{\natexlab{a}}.

\bibitem[Wang et~al.(2023{\natexlab{a}})Wang, Chen, Pei, Xie, Kang, Zhang, Xu, Xiong, Dutta, Schaeffer, et~al.]{wang2023decodingtrust}
Wang, B., Chen, W., Pei, H., Xie, C., Kang, M., Zhang, C., Xu, C., Xiong, Z., Dutta, R., Schaeffer, R., et~al.
\newblock Decodingtrust: A comprehensive assessment of trustworthiness in gpt models.
\newblock \emph{arXiv preprint arXiv:2306.11698}, 2023{\natexlab{a}}.

\bibitem[Wang et~al.(2020)Wang, Tang, Ma, Wu, Popuri, Okhonko, and Pino]{wang2020fairseq}
Wang, C., Tang, Y., Ma, X., Wu, A., Popuri, S., Okhonko, D., and Pino, J.
\newblock Fairseq s2t: Fast speech-to-text modeling with fairseq.
\newblock \emph{arXiv preprint arXiv:2010.05171}, 2020.

\bibitem[Wang et~al.(2022)Wang, Yang, Hu, Li, Lin, Gan, Liu, Liu, and Wang]{wang2022git}
Wang, J., Yang, Z., Hu, X., Li, L., Lin, K., Gan, Z., Liu, Z., Liu, C., and Wang, L.
\newblock Git: A generative image-to-text transformer for vision and language.
\newblock \emph{arXiv preprint arXiv:2205.14100}, 2022.

\bibitem[Wang et~al.(2023{\natexlab{b}})Wang, Lv, Yu, Hong, Qi, Wang, Ji, Yang, Zhao, Song, Xu, Xu, Li, Dong, Ding, and Tang]{wang2023cogvlm}
Wang, W., Lv, Q., Yu, W., Hong, W., Qi, J., Wang, Y., Ji, J., Yang, Z., Zhao, L., Song, X., Xu, J., Xu, B., Li, J., Dong, Y., Ding, M., and Tang, J.
\newblock Cogvlm: Visual expert for pretrained language models, 2023{\natexlab{b}}.

\bibitem[Wang et~al.(2021{\natexlab{b}})Wang, Liu, Gui, Zhang, Zou, Zhou, Ye, Zhang, Zheng, Pang, et~al.]{wang2021textflint}
Wang, X., Liu, Q., Gui, T., Zhang, Q., Zou, Y., Zhou, X., Ye, J., Zhang, Y., Zheng, R., Pang, Z., et~al.
\newblock Textflint: Unified multilingual robustness evaluation toolkit for natural language processing.
\newblock In \emph{Proceedings of the 59th Annual Meeting of the Association for Computational Linguistics and the 11th International Joint Conference on Natural Language Processing: System Demonstrations}, pp.\  347--355, 2021{\natexlab{b}}.

\bibitem[Watanabe et~al.(2018)Watanabe, Hori, Karita, Hayashi, Nishitoba, Unno, {Enrique Yalta Soplin}, Heymann, Wiesner, Chen, Renduchintala, and Ochiai]{watanabe2018espnet}
Watanabe, S., Hori, T., Karita, S., Hayashi, T., Nishitoba, J., Unno, Y., {Enrique Yalta Soplin}, N., Heymann, J., Wiesner, M., Chen, N., Renduchintala, A., and Ochiai, T.
\newblock {ESPnet}: End-to-end speech processing toolkit.
\newblock In \emph{Proceedings of Interspeech}, pp.\  2207--2211, 2018.
\newblock \doi{10.21437/Interspeech.2018-1456}.
\newblock URL \url{http://dx.doi.org/10.21437/Interspeech.2018-1456}.

\bibitem[Xu et~al.(2021)Xu, Baevski, Likhomanenko, Tomasello, Conneau, Collobert, Synnaeve, and Auli]{xu2021self}
Xu, Q., Baevski, A., Likhomanenko, T., Tomasello, P., Conneau, A., Collobert, R., Synnaeve, G., and Auli, M.
\newblock Self-training and pre-training are complementary for speech recognition.
\newblock In \emph{ICASSP 2021-2021 IEEE International Conference on Acoustics, Speech and Signal Processing (ICASSP)}, pp.\  3030--3034. IEEE, 2021.

\bibitem[Ye et~al.(2023{\natexlab{a}})Ye, Xu, Xu, Ye, Yan, Zhou, Wang, Hu, Shi, Shi, et~al.]{ye2023mplugv1}
Ye, Q., Xu, H., Xu, G., Ye, J., Yan, M., Zhou, Y., Wang, J., Hu, A., Shi, P., Shi, Y., et~al.
\newblock mplug-owl: Modularization empowers large language models with multimodality.
\newblock \emph{arXiv preprint arXiv:2304.14178}, 2023{\natexlab{a}}.

\bibitem[Ye et~al.(2023{\natexlab{b}})Ye, Xu, Ye, Yan, Liu, Qian, Zhang, Huang, and Zhou]{ye2023mplug}
Ye, Q., Xu, H., Ye, J., Yan, M., Liu, H., Qian, Q., Zhang, J., Huang, F., and Zhou, J.
\newblock mplug-owl2: Revolutionizing multi-modal large language model with modality collaboration.
\newblock \emph{arXiv preprint arXiv:2311.04257}, 2023{\natexlab{b}}.

\bibitem[Zeng et~al.(2023)Zeng, Liu, Du, Wang, Lai, Ding, Yang, Xu, Zheng, Xia, Tam, Ma, Xue, Zhai, Chen, Liu, Zhang, Dong, and Tang]{zeng2023glm-130b}
Zeng, A., Liu, X., Du, Z., Wang, Z., Lai, H., Ding, M., Yang, Z., Xu, Y., Zheng, W., Xia, X., Tam, W.~L., Ma, Z., Xue, Y., Zhai, J., Chen, W., Liu, Z., Zhang, P., Dong, Y., and Tang, J.
\newblock {GLM}-130b: An open bilingual pre-trained model.
\newblock In \emph{The Eleventh International Conference on Learning Representations (ICLR)}, 2023.
\newblock URL \url{https://openreview.net/forum?id=-Aw0rrrPUF}.

\bibitem[Zhang et~al.(2022)Zhang, Roller, Goyal, Artetxe, Chen, Chen, Dewan, Diab, Li, Lin, Mihaylov, Ott, Shleifer, Shuster, Simig, Koura, Sridhar, Wang, and Zettlemoyer]{zhang2022opt}
Zhang, S., Roller, S., Goyal, N., Artetxe, M., Chen, M., Chen, S., Dewan, C., Diab, M., Li, X., Lin, X.~V., Mihaylov, T., Ott, M., Shleifer, S., Shuster, K., Simig, D., Koura, P.~S., Sridhar, A., Wang, T., and Zettlemoyer, L.
\newblock Opt: Open pre-trained transformer language models, 2022.

\bibitem[Zhao et~al.(2022)Zhao, Yu, Ma, Yu, Mei, Wang, He, Yuille, and Kortylewski]{zhao2022ood}
Zhao, B., Yu, S., Ma, W., Yu, M., Mei, S., Wang, A., He, J., Yuille, A., and Kortylewski, A.
\newblock Ood-cv: a benchmark for robustness to out-of-distribution shifts of individual nuisances in natural images.
\newblock In \emph{European Conference on Computer Vision}, pp.\  163--180. Springer, 2022.

\bibitem[Zhao et~al.(2023)Zhao, Pang, Du, Yang, Li, Cheung, and Lin]{zhao2023evaluate}
Zhao, Y., Pang, T., Du, C., Yang, X., Li, C., Cheung, N.-M., and Lin, M.
\newblock On evaluating adversarial robustness of large vision-language models.
\newblock In \emph{Advances in Neural Information Processing Systems (NeurIPS)}, 2023.

\bibitem[Zheng et~al.(2023)Zheng, Chiang, Sheng, Zhuang, Wu, Zhuang, Lin, Li, Li, Xing, et~al.]{zheng2023judging}
Zheng, L., Chiang, W.-L., Sheng, Y., Zhuang, S., Wu, Z., Zhuang, Y., Lin, Z., Li, Z., Li, D., Xing, E., et~al.
\newblock Judging llm-as-a-judge with mt-bench and chatbot arena.
\newblock \emph{arXiv preprint arXiv:2306.05685}, 2023.

\bibitem[Zhou et~al.(2021)Zhou, Zhang, Chen, Li, Tensmeyer, Yu, Gu, Xu, and Sun]{zhou2021lafite}
Zhou, Y., Zhang, R., Chen, C., Li, C., Tensmeyer, C., Yu, T., Gu, J., Xu, J., and Sun, T.
\newblock Lafite: Towards language-free training for text-to-image generation.
\newblock \emph{arXiv preprint arXiv:2111.13792}, 2021.

\bibitem[Zhu et~al.(2023{\natexlab{a}})Zhu, Chen, Shen, Li, and Elhoseiny]{zhu2023minigpt}
Zhu, D., Chen, J., Shen, X., Li, X., and Elhoseiny, M.
\newblock Minigpt-4: Enhancing vision-language understanding with advanced large language models.
\newblock \emph{arXiv preprint arXiv:2304.10592}, 2023{\natexlab{a}}.

\bibitem[Zhu et~al.(2023{\natexlab{b}})Zhu, Wang, Zhou, Wang, Chen, Wang, Yang, Ye, Gong, Zhang, et~al.]{zhu2023promptbench}
Zhu, K., Wang, J., Zhou, J., Wang, Z., Chen, H., Wang, Y., Yang, L., Ye, W., Gong, N.~Z., Zhang, Y., et~al.
\newblock Promptbench: Towards evaluating the robustness of large language models on adversarial prompts.
\newblock \emph{arXiv preprint arXiv:2306.04528}, 2023{\natexlab{b}}.

\end{thebibliography}
\bibliographystyle{icml2023}

%%%%%%%%%%%%%%%%%%%%%%%%%%%%%%%%%%%%%%%%%%%%%%%%%%%%%%%%%%%%%%%%%%%%%%%%%%%%%%%
%%%%%%%%%%%%%%%%%%%%%%%%%%%%%%%%%%%%%%%%%%%%%%%%%%%%%%%%%%%%%%%%%%%%%%%%%%%%%%%
% APPENDIX
%%%%%%%%%%%%%%%%%%%%%%%%%%%%%%%%%%%%%%%%%%%%%%%%%%%%%%%%%%%%%%%%%%%%%%%%%%%%%%%
%%%%%%%%%%%%%%%%%%%%%%%%%%%%%%%%%%%%%%%%%%%%%%%%%%%%%%%%%%%%%%%%%%%%%%%%%%%%%%%
\newpage
\appendix
\onecolumn
\section{Corrupted Examples}

\subsection{Examples of Text Corruptions}
\label{apx:tti_example}

Examples of the 23 distinct text corruptions are presented in~\Cref{tab:text2image_example}.

\subsection{Examples of Image Corruptions}
\label{apx:itt_example}
\begin{figure*}[!t] 
  \centering
  \includegraphics[width=0.85\textwidth]{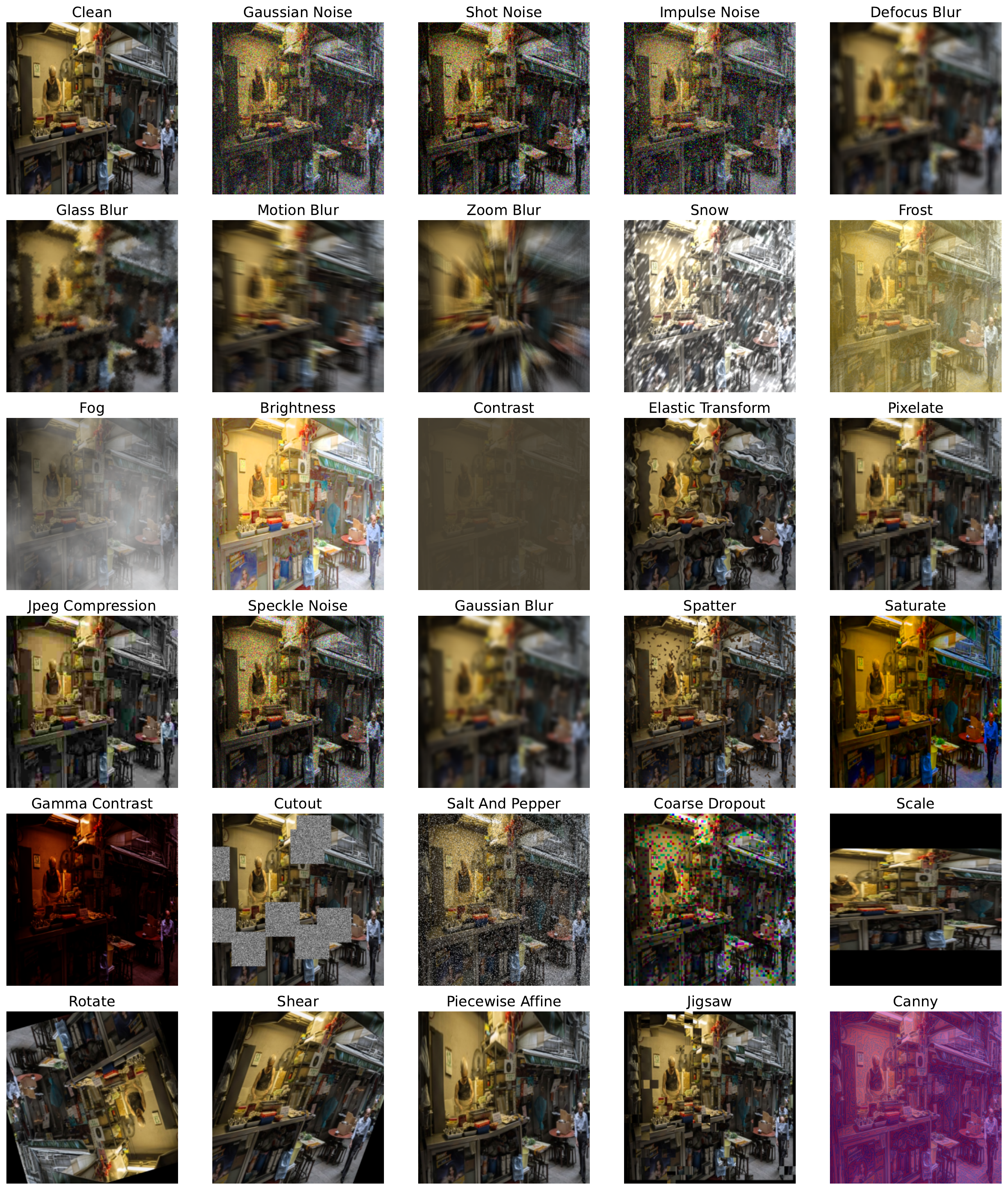}
    \caption{Examples of 29 distinct image corruptions applied to a single image.}
  \label{fig:image2text_example}
\end{figure*}

Examples of the 29 distinct image corruptions are illustrated in~\Cref{fig:image2text_example}.

\section{Additional Experimental Results}
\subsection{Text to Image}
\label{apx:tti_uni}
The inconsistency score, calculated based on unimodality, is presented in~\Cref{tab:text2image_uni}. This score is derived from the comparison between the image generated from the clean input text and the images generated from the corrupted texts. The image similarity calculations are performed using the image encoder from the CLIP model~\footnote{\scriptsize\url{https://huggingface.co/laion/CLIP-ViT-L-14-laion2B-s32B-b82K}}. We observe that the score for various models on our selected challenging dataset is significantly lower than that on randomly sampled data, which further demonstrate the effectiveness of our selection strategy. However, one limitation of this metric is that a model consistently yielding subpar results may still attain a high consistency score. For instance, despite its inferior quality in correlating the original caption with its generated images, the model ranked first, Lafite, manages to secure a high similarity score between its generated images. Additionally, it is noteworthy that the high-quality results of the IF model are contingent upon extensive computation, necessitating prolonged image generation times. In contrast, the LCM model, which operates with greater speed, manages to achieve a comparable level of consistency performance.

\subsection{Image to Text}
\label{apx:itt_uni}
The inconsistency score, based on unimodality, is presented in~\Cref{tab:image2text_uni}. This metric is derived from the semantic similarity between captions generated for clean input images and those generated for corrupted images. The similarity computations utilize the text encoder from the CLIP model referenced in~\Cref{apx:tti_uni}. The results show that models perform considerably better on our meticulously chosen challenging data than on randomly selected data, thereby confirming the efficacy of our selection strategy. Furthermore, it is evident that MLLMs still significantly outperform non-LLM-based models when dealing with corrupted image inputs, showcasing the robustness of LLMs in resisting the impact of such distortions.

\begin{table*}[!b]
\centering
\caption{Examples of our $23$ distinct text corruptions categorized into three levels: character level, word level, and sentence level.}
\vspace{0.2cm}
\small
\begin{tabular}{c|m{5cm}|m{8cm}}
\toprule
\hline
\textbf{Level} & \centering\arraybackslash\textbf{Perturbation Type} &  \centering\arraybackslash\textbf{Caption} \\
\hline
- & \centering clean & The band, including one man in black jacket and two women standing next to blue corrugated walls. \\
\hline
\multirow{20}{*}{Char} & \centering Substitute Char by Ocr & The band, inclodin\diff{9} one man in b\diff{1}ack jacket and two women standing next t\diff{u} blue corrugated walls. \\
\cline{2-3}
& \centering Substitute Char by Keyboard & The band, including one man in black jacket and two \diff{Q}omen \diff{z}randing n\diff{3}xt to blue cotrugat\diff{D}d walls. \\
\cline{2-3}
& \centering Insert Char Randomly & The ba\diff{r}nd, including one man in black \diff{l}jack\diff{d}et and two wom\diff{3}en standing next to blue corru\diff{9}gat\diff{E}ed walls. \\
\cline{2-3}
& \centering Substitute Char Randomly & The band, imclu\diff{(}ing one man in black j\diff{y}c\diff{O}et and two women stan\diff{x}in\diff{b} next to blue corrugated w\diff{\&}lls. \\
\cline{2-3}
& \centering Swap Char Randomly & The band, including one man in \diff{balck} jacket and two women standing \diff{enxt} to blue \diff{corruagtde} \diff{walsl}. \\
\cline{2-3}
& \centering Delete Char Randomly & The \diff{bad}, including one man in black jacket and two \diff{woem} \diff{stadig} next to \diff{blu} corrugated walls. \\
\cline{2-3}
& \centering Uppercase Char Randomly & Th\diff{E} ban\diff{D}, inc\diff{L}ud\diff{I}ng on\diff{E} \diff{M}a\diff{N} in \diff{B}l\diff{A}ck jacket and two w\diff{O}\diff{M}\diff{E}n standing n\diff{E}xt to blu\diff{E} c\diff{O}rr\diff{U}gat\diff{E}d walls. \\
\cline{2-3}
& \centering Repeat Characters & \diff{TThhee  bbaanndd,,  iinncclluuddiinngg  oonnee  mmaann  iinn  bbllaacckk  jjaacckkeett  ...} \\
\cline{2-3}
& \centering Leet Letters & The band, including on\diff{3} \diff{3}an in black jack\diff{3}t and tw\diff{0}  \diff{3}o\diff{33}n standing next t\diff{0} \diff{6}lu\diff{3} corrugated \diff{3}alls. \\
\cline{2-3}
& \centering Whitespace Perturbation & The band, including one \diff{ma n} in black \diff{ \  jack et} and  two \diff{w omen}  standing next to \diff{b lue} corrugated walls. \\
\cline{2-3}
& \centering Substitute with Homoglyphs & The b$\mathfrak{\diff{a}}$nd, inclu$\mathrm{\diff{d}}$ing one $\mathbf{\diff{m}}$an in b$\mathrm{\diff{l}}$ack jac$\mathtt{\diff{k}}$et and two wom$\mathsf{\diff{e}}$n standing ne$\mathnormal{\diff{x}}$t to blue corru$\mathfrak{\diff{g}}$ated walls. \\
\hline
\multirow{15}{*}{Word} & \centering Synonym Replacement & The band, \diff{admit unmatchable homo in pitch dark jacket crown} and \diff{deuce char brook} adjacent to blue \diff{sky corrugate} walls. \\
\cline{2-3}
& \centering Random Deletion & \diff{[The]} band, including one man in black jacket \diff{[and two]} women standing next \diff{[to]} blue corrugated walls. \\
\cline{2-3}
& \centering Random Swap & The band, \diff{man and including in black jacket one} two women standing next to blue corrugated walls. \\
\cline{2-3}
& \centering Random Insertion & The band, \diff{let in} including one \diff{succeeding} man in black jacket and two women standing next to blue corrugated walls. \\
\cline{2-3}
& \centering Misspell Word & The \diff{banda}, including \diff{onw} man \diff{In} black jacket and two women standing \diff{nexto} to blue corrugated walls. \\
\cline{2-3}
& \centering Abbreviate Word & The band , including one man in \diff{blk} jacket and two women standing next \diff{tuh blu} corrugated walls . \\
\cline{2-3}
& \centering Multilingual Dictionary Based \\ Code Switch & The band, \diff{compresa} one man in black \diff{jaqueta} and two women standing next to blue corrugated walls. \\
\cline{2-3}
& \centering Close Homophones Swap & \diff{Thee Banned}, including one man \diff{Inn} black \diff{Jackett} and \diff{Thuy} women standing \diff{Nex} to \diff{Blew} corrugated \diff{Wall'S}. \\
\hline
\multirow{7}{*}{Sentence} & \centering CheckList & \diff{@YKIW} The band, including one man in black jacket and two women standing next to blue corrugated \diff{https://t.co/DdIQ} walls. \\
\cline{2-3}
& \centering Back Translation & The band, \diff{which includes a man in a black jacket and two women, stands} next to blue corrugated \diff{iron} walls. \\
\cline{2-3}
& \centering Style Paraphraser & The band, \diff{with one man in black and two women in blue corrugated walls, stands in the street}. \\
\cline{2-3}
& \centering Paraphrase & \diff{One man was in a black jacket, and two women were standing next to blue corrugated walls as members of the band}. \\
\hline
\bottomrule
\end{tabular}
\label{tab:text2image_example}
\end{table*}
\begin{table*}[!t]
\centering
\caption{Performance comparison of various \textbf{text-to-image} models evaluated by self-consistency scores (uni-modality) across different corruption intensities and data selection levels. Scores represent the average multiplied cosine similarities (max 2300) between the images generated from original captions and the generated images for the captions under different corruption conditions.}
\vspace{0.2cm}
\label{tab:text2image_uni}
\begin{tabular*}{0.85\textwidth}{@{\extracolsep{\fill}}lcccccc}
\toprule
\multirow{2}{*}{Models} & \multicolumn{2}{c}{Hard} & \multicolumn{2}{c}{Random} & \multirow{2}{*}{Average} \\  \cmidrule(lr){2-3} \cmidrule(lr){4-5}
 & Heavy & Light & Heavy & Light & \\
\midrule
Lafite~\cite{zhou2021lafite} & 1697 & 1790 & 1679 & 1783 & 1737.25 \\
IF XL v1.0~\cite{deepfloyd_if_2023} & 1444 & 1615 & 1533 & 1732 & 1581.00 \\
IF L v1.0~\cite{deepfloyd_if_2023} & 1414 & 1581 & 1529 & 1732 & 1564.00 \\
SDXL Turbo~\cite{sauer2023adversarial} & 1412 & 1614 & 1474 & 1697 & 1549.25 \\
IF M v1.0~\cite{deepfloyd_if_2023}   & 1388 & 1549 & 1522 & 1725 & 1546.00 \\
Kandinsky 3~\cite{arkhipkin2023kandinsky} & 1377 & 1573 & 1478 & 1706 & 1533.50 \\
LCM (Dreamshaper v7)~\cite{luo2023latent, lykon_dreamshaper7_2023} & 1421 & 1560 & 1483 & 1646 & 1527.50 \\
LCM (SSD-1B)~\cite{luo2023latent, gupta2024progressive} & 1388 & 1547 & 1463 & 1649 & 1511.75 \\
Stable Diffusion Turbo~\cite{sauer2023adversarial} & 1384 & 1564 & 1418 & 1630 & 1499.00 \\
SSD 1B~\cite{gupta2024progressive} & 1368 & 1518 & 1441 & 1621 & 1487.00 \\
LCM LoRA (SSD-1B)~\cite{luo2023lcm, gupta2024progressive} & 1363 & 1522 & 1438 & 1621 & 1486.00 \\
LCM LoRA (SDXL)~\cite{luo2023lcm, podell2023sdxl} & 1328 & 1503 & 1394 & 1586 & 1452.75 \\
Small Stable Diffusion~\cite{kim2023architectural} & 1344 & 1488 & 1393 & 1572 & 1449.25 \\
LCM (SDXL)~\cite{luo2023latent, podell2023sdxl} & 1307 & 1475 & 1398 & 1595 & 1443.75 \\
SDXL Refiner~\cite{podell2023sdxl} & 1305 & 1482 & 1378 & 1572 & 1434.25 \\
Kandinsky 2.2~\cite{kandinsky2}          & 1290 & 1454 & 1389 & 1598 & 1432.75 \\
SDXL Base~\cite{podell2023sdxl}   & 1307 & 1476 & 1375 & 1568 & 1431.50 \\
Dreamlike Photoreal 2.0~\cite{dreamlike_art_2023} & 1279 & 1445 & 1370 & 1560 & 1413.50 \\
LCM LoRA (SD v1)~\cite{luo2023lcm, rombach2022high} & 1276 & 1416 & 1395 & 1566 & 1413.25 \\
Dreamshaper v7~\cite{lykon_dreamshaper7_2023} & 1272 & 1419 & 1397 & 1562 & 1412.50 \\
Openjourney v4~\cite{prompthero_openjourneyv4_2023} & 1258 & 1413 & 1340 & 1517 & 1382.00 \\
Stable Diffusion v2~\cite{rombach2022high}        & 1211 & 1380 & 1316 & 1516 & 1355.75 \\
Anything Midjourney v4.1~\cite{prompthero_openjourneyv4_2023}     & 1225 & 1385 & 1313 & 1492 & 1353.75 \\
Glide~\cite{nichol2022glide} & 1278 & 1342 & 1326 & 1420 & 1341.50 \\
Stable Diffusion v1~\cite{rombach2022high} & 1217 & 1364 & 1300 & 1479 & 1340.00 \\
Unidiffuser~\cite{bao2023one} & 1168 & 1298 & 1293 & 1461 & 1305.00 \\
Karlo-v1.0.alpha~\cite{kakaobrain2022karlo-v1-alpha}                 & 1123 & 1333 & 1220 & 1476 & 1288.00 \\
\bottomrule
\end{tabular*}
\end{table*}
\begin{table*}[!t]
\centering
\caption{Performance comparison of various \textbf{image-to-text} models evaluated by self-consistency scores (uni-modality) across different corruption intensities and data selection levels. Scores represent the average multiplied cosine similarities (max 2900) between the captions generated for the original images and the captions generated for the images under different corruption conditions.}
\vspace{0.2cm}
\label{tab:image2text_uni}
\begin{tabular*}{1.0\textwidth}{@{\extracolsep{\fill}}lccccccc}
\toprule
\multirow{2}{*}{Models} & \multirow{2}{*}{LLM} &\multicolumn{2}{c}{Hard} & \multicolumn{2}{c}{Random} & \multirow{2}{*}{Average} \\  \cmidrule(lr){3-4} \cmidrule(lr){5-6} 
&  & Heavy & Light & Heavy & Light & \\
\midrule
\first InstructBLIP~\cite{dai2023instructblip} & Vicuna 13B~\cite{zheng2023judging} & 2017 & 2254 & 2238 & 2389 & 2224.50 \\
\second LLaVA-v1.5~\cite{liu2023improved} & Vicuna 13B~\cite{zheng2023judging} & 1956 & 2192 & 2275 & 2404 & 2206.75 \\
\third InstructBLIP~\cite{dai2023instructblip} & Vicuna 7B~\cite{zheng2023judging} & 1980 & 2223 & 2229 & 2386 & 2204.50 \\
InstructBLIP~\cite{dai2023instructblip} & Flan T5 XL~\cite{chung2022scaling}  & 1945 & 2241 & 2200 & 2344 & 2182.50 \\
LLaVA-v1.5~\cite{liu2023improved} & Vicuna 7B~\cite{zheng2023judging} & 1918 & 2160 & 2249 & 2388 & 2178.75 \\
BLIP2~\cite{li2023blip} & Flan T5 XXL~\cite{chung2022scaling} & 1819 & 2123 & 2215 & 2399 & 2139.00 \\
LLaVA ~\cite{liu2023visual}  & Vicuna 13B~\cite{zheng2023judging}  & 1885 & 2122 & 2203 & 2337 & 2136.75 \\
BLIP2~\cite{li2023blip} & Flan T5 XL~\cite{chung2022scaling} & 1777 & 2131 & 2201 & 2427 & 2134.00 \\
InstructBLIP~\cite{dai2023instructblip} & Flan T5 XXL~\cite{chung2022scaling} & 1966 & 2172 & 2125 & 2251 & 2128.50 \\
LLaVA~\cite{liu2023improved} & LLaMA2 13B~\cite{touvron2023llama2}     & 1838 & 2080 & 2182 & 2331 & 2107.75 \\
BLIP2~\cite{li2023blip} &  OPT-6.7b~\cite{zhang2022opt} & 1776 & 2088 & 2141 & 2332 & 2084.25 \\
mPLUG-Owl~\cite{ye2023mplugv1}  & LLaMA 7B~\cite{touvron2023llama} & 1842 & 2107 & 2107 & 2274 & 2082.50 \\
LLaVA~\cite{liu2023improved} & MPT 7B~\cite{MosaicML2023Introducing} & 1827 & 2056 & 2141 & 2288 & 2078.00 \\
LLaMA-Adapter v2~\cite{gao2023llamaadapterv2}  & LLaMA 7B~\cite{touvron2023llama}     & 1840 & 2059 & 2133 & 2264 & 2074.00 \\
mPLUG-Owl2~\cite{ye2023mplug}  & LLaMA2 7B~\cite{touvron2023llama2} & 1748 & 2054 & 2104 & 2324 & 2057.50 \\
BLIP2~\cite{li2023blip} & OPT-2.7b~\cite{zhang2022opt} & 1714 & 2030 & 2084 & 2284 & 2028.00 \\
LLaVA~\cite{liu2023improved} & LLaMA2 7B~\cite{touvron2023llama2}     & 1769 & 1997 & 2102 & 2243 & 2027.75 \\
LaVIN~\cite{luo2023towards,luo2023cheap} & LLaMA 13B~\cite{touvron2023llama}      & 1759 & 2036 & 2047 & 2249 & 2022.75 \\
mPLUG-Owl (multilingual)~\cite{ye2023mplugv1}  & LLaMA 7B~\cite{touvron2023llama} & 1765 & 2008 & 2071 & 2229 & 2018.25 \\
ShareGPT4V~\cite{chen2023sharegpt4v} & Vicuna 7B~\cite{zheng2023judging} & 1699 & 1956 & 2105 & 2277 & 2009.25 \\
CogVLM~\cite{wang2023cogvlm,hong2023cogagent} & Vicuna 7B~\cite{zheng2023judging} & 1819 & 2093 & 1937 & 2119 & 1992.00 \\
MiniGPT-4~\cite{zhu2023minigpt} & Vicuna 13B~\cite{zheng2023judging}                    & 1765 & 1938 & 2052 & 2150 & 1976.25 \\
MiniGPT-4~\cite{zhu2023minigpt} & Vicuna 7B~\cite{zheng2023judging} & 1759 & 1936 & 1999 & 2089 & 1945.75 \\
MiniGPT-4~\cite{zhu2023minigpt} & LLaMA2 7B~\cite{touvron2023llama2} & 1722 & 1875 & 2015 & 2084 & 1924.00 \\
ImageBind-LLM~\cite{han2023imagebind} & Open Chinese LLaMA 7B~\cite{openlmlab, touvron2023llama} & 1585 & 1833 & 2005 & 2165 & 1897.00 \\
BLIP Large~\cite{li2022blip} & - & 1487 & 1805 & 1983 & 2238 & 1878.25 \\
VisualGLM~\cite{du2022glm,ding2021cogview} & ChatGLM-6B~\cite{du2022glm, zeng2023glm-130b} & 1598 & 1811 & 1849 & 1953 & 1802.75 \\
Unidiffuser~\cite{bao2023one} & - & 1484 & 1669 & 1894 & 2085 & 1783.00 \\
ViT-GPT2~\cite{nlp_connect_2022} & GPT2~\cite{radford2019language} & 1262 & 1468 & 1918 & 2169 & 1704.25 \\
GIT Large~\cite{wang2022git} & - & 1336 & 1556 & 1878 & 2046 & 1704.00 \\
BLIP Base~\cite{li2022blip} & - & 1192 & 1410 & 1887 & 2153 & 1660.50 \\
Qwen-VL-Chat~\cite{Qwen-VL}& Qwen-7B~\cite{bai2023qwen} & 1317 & 1611 & 1696 & 1901 & 1631.25 \\
OpenFlamingo~\cite{awadalla2023openflamingo,Alayrac2022FlamingoAV} & MPT 7B~\cite{MosaicML2023Introducing} & 1506 & 1681 & 1530 & 1708 & 1606.25 \\
Multimodal-GPT~\cite{gong2023multimodalgpt} & LLaMA 7B~\cite{touvron2023llama}    & 1400 & 1591 & 1639 & 1771 & 1600.25 \\
GIT Base~\cite{wang2022git} & - & 1232 & 1394 & 1769 & 1965 & 1590.00 \\
OpenFlamingo~\cite{awadalla2023openflamingo,Alayrac2022FlamingoAV} & RedPajama 3B~\cite{redPajama3b} & 1367 & 1537 & 1488 & 1660 & 1513.00 \\
OpenFlamingo~\cite{awadalla2023openflamingo,Alayrac2022FlamingoAV} & MPT 1B~\cite{mpt1b} & 1219 & 1404 & 1312 & 1519 & 1363.50 \\
% Pix2Struct Base~\cite{lee2023pix2struct} & - & 1302 & 1429 & 1230 & 1399 & 1340.00 \\
MiniGPT v2~\cite{chen2023minigpt} & LLaMA2 7B~\cite{touvron2023llama2} & 1131 & 1204 & 1318 & 1381 & 1258.50 \\
% Pix2Struct Large~\cite{lee2023pix2struct} & - & 1070 & 1243 & 1041 & 1224 & 1144.50\\
\bottomrule
\end{tabular*}
\end{table*}

\end{document}